\documentclass[runningheads]{llncs}

\usepackage[T1]{fontenc}    
\usepackage{hyperref}       
\usepackage{url}            
\usepackage{booktabs}       
\usepackage{amsfonts}       
\usepackage{nicefrac}       
\usepackage{microtype}      
\usepackage{lipsum}
\usepackage{graphicx}       
\usepackage{amsmath,amsfonts}
\usepackage{algorithmic}
\usepackage{algorithm}
\usepackage{array}
\usepackage{cuted}
\usepackage{textcomp}
\usepackage{stfloats}
\usepackage{verbatim}
\usepackage{epsfig}
\usepackage{amssymb}
\usepackage{subcaption}
\usepackage{multirow}

\usepackage{pifont}
\usepackage{float}
\usepackage{breqn}
\usepackage[normalem]{ulem} 

\newcommand{\cmark}{\ding{51}}%
\newcommand{\xmark}{\ding{55}}%
\newcommand{\METHOD}{FLAC}
\newcolumntype{?}{!{\vrule width 1.5pt}}
\begin{document}

\title{FLAC: Fairness-Aware Representation Learning by Suppressing Attribute-Class Associations}
\titlerunning{Accepted for publication at IEEE TPAMI}
\author{Ioannis Sarridis\inst{1,2} \and
Christos Koutlis\inst{1} \and
Symeon Papadopoulos\inst{1} \and
Christos Diou\inst{2}}
\authorrunning{I. Sarridis et al.}
\institute{Information Technologies Institute, Centre for Research and Technology Hellas, 6th km Charilaou-Thermi Rd, Thessaloniki, 57001, Greece
\email{\{gsarridis,ckoutlis,papadop\}@iti.gr}\\
\and
Department of Informatics and Telematics, Harokopio University of Athens Omirou 9, Tavros, 17778, Attika, Greece\\
\email{cdiou@hua.gr}
}
\maketitle

\begin{abstract}
Bias in computer vision systems can perpetuate or even amplify discrimination against certain populations. Considering that bias is often introduced by biased visual datasets,  many recent research efforts focus on training fair models using such data. However, most of them heavily rely on the availability of protected attribute labels in the dataset, which limits their applicability, while label-unaware approaches, i.e., approaches operating without such labels, exhibit considerably lower performance. 
To overcome these limitations, this work introduces FLAC, a methodology that minimizes mutual information between the features extracted by the model and a protected attribute, without the use of attribute labels. To do that, FLAC proposes a sampling strategy that highlights underrepresented samples in the dataset, and casts the problem of learning fair representations as a probability matching problem that leverages representations extracted by a bias-capturing classifier. It is theoretically shown that FLAC can indeed lead to fair representations, that are independent of the protected attributes.
FLAC surpasses the current state-of-the-art on Biased-MNIST, CelebA, and UTKFace, by 29.1\%,  18.1\%, and 21.9\%, respectively. Additionally,  FLAC exhibits 2.2\% increased accuracy on ImageNet-A and up to 4.2\% increased accuracy on Corrupted-Cifar10. Finally, in most experiments, FLAC even outperforms the bias label-aware  state-of-the-art methods.
\end{abstract}
\section{Introduction}

\begin{figure}[h]
    \centering

        \includegraphics[width=1.1\linewidth]{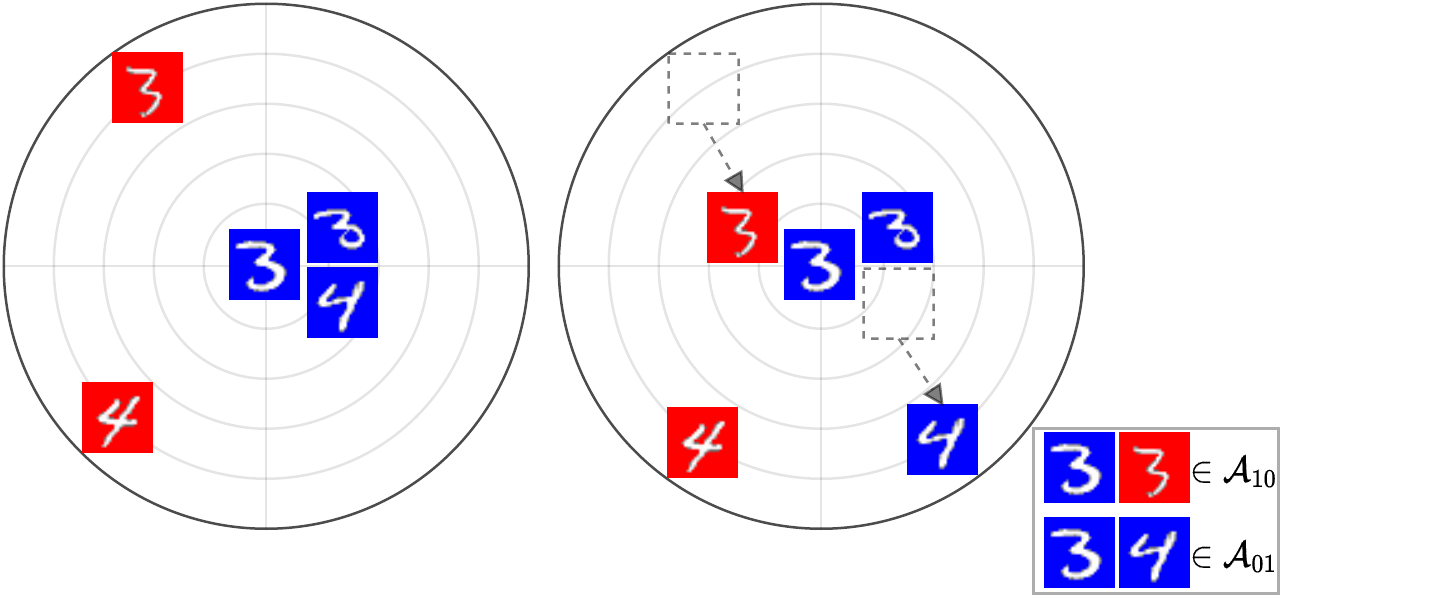}
    \caption{ 
     Distances between the central sample and the rest of the samples belonging to Biased-MNIST, a dataset that demonstrates a strong association between the target labels (i.e., \textit{digits}) and the protected attributes (i.e., \textit{colors}). Subfigure on the left shows that using the standard task-specific loss introduces bias to the representations which \METHOD, the proposed approach, can effectively mitigate as shown on the right. In particular, \METHOD\ focuses on situations where task-specific losses are susceptible to data bias, namely pairs with the same target label but different protected attribute values (i.e., $\mathcal{A}_{10}$) and pairs with different target labels and the same protected attribute value (i.e., $\mathcal{A}_{01}$).}
    \label{fig:fig1}
\end{figure}
During the past decade, AI models have demonstrated exceptional performance in a growing number of application areas; however, there have also been numerous incidents where failures in the AI systems have disproportionately affected certain individuals or groups of people \cite{barocasfairml}.
Bias in Artificial Intelligence (AI) typically refers to AI systems that demonstrate discriminatory behavior (e.g., high errors) against groups or populations w.r.t. certain \textit{protected} or \textit{sensitive} attributes that in several jurisdictions are legally protected from discrimination, e.g., race, gender, age, and religion. This makes fairness and non-discrimination an important ethical and often legal requirement for AI models, and a prerequisite for their wide adoption and use in many real-world applications \cite{taigman2014deepface,creswell2018generative,bobadilla2013recommender,tan2020efficientdet}.

Bias can emerge in several Computer Vision (CV) systems, often in applications involving the processing of face images but also in other domains where visual data is involved \cite{fabbrizzi2022survey}. Many research efforts focusing on face verification fairness \cite{frisella2022quantifying,gluge2020not,cavazos2020accuracy} highlight that AI models tend to be inaccurate for black people, especially for black women \cite{sixta2020fairface}. In addition, similar biased behavior against under-represented groups has been noticed in AI systems for face recognition \cite{sixta2020fairface}, facial expression recognition \cite{xu2020investigating}, and facial attribute analysis \cite{zheng2020survey}.

Taking into account the importance of fairness in AI, several approaches for learning fair representations have been proposed in recent years. 
The underlying idea of such approaches is that if the representations of samples belonging to different groups w.r.t. a protected attribute are similar, then the classifier built on top of them will inevitably make decisions irrespective of the group to which the samples belong. This serves as a means for achieving demographic parity (statistical parity) between the groups of different protected attribute values \cite{zhao2022inherent}. 
Many approaches make use of the protected attribute labels (also known as bias labels in the literature) provided in the training set \cite{kim2019lnl,tartaglione2021end,hong2021bb,barbano2022fairkl}. We refer to those as Bias Label-Aware (BL-A) methods. Despite the effectiveness of such methods in terms of bias mitigation, they can only be applied in a narrow range of problems and datasets, due to their reliance on the availability of protected attribute labels. To overcome this limitation, several recent efforts employ the representations derived by a bias-capturing classifier for mitigating the bias without being aware of the protected attribute labels \cite{clark2019LM,cadene2019rubi,bahng2020rebias,nam2020LfF}. We refer to these methods as Bias Label-Unaware (BL-U). The bias-capturing classifier is a model that attempts to encapsulate information for the protected attribute. In addition, BL-U approaches can be applied in cases where bias is not categorical. For instance, representations with texture bias can be derived by a CNN with small receptive fields \cite{bahng2020rebias}. However, the performance of BL-U approaches has been shown to lag considerably behind the one of BL-A methods \cite{barbano2022fairkl,hong2021bb}. 

Mutual Information (MI) can be used to measure the dependence between the representation provided by the main model and the protected attribute values. By minimizing MI during representation learning, one can deter the main model from using a given protected attribute to predict the label. 
In this way, fairness can be achieved through a group-invariant representation, which is known to guarantee accuracy parity between the groups defined by the protected attribute \cite{zhao2022inherent}. However, solving the above MI minimization problem would require access to the protected attributes (i.e., would need a BL-A method).
In this paper we introduce \METHOD, a BL-U approach that leverages the representations of a bias-capturing classifier to force an initial potentially biased model to learn fairer representations. In particular, we cast the reduction of the MI minimization problem into a simpler probability matching problem between the similarity distributions of the main model and the bias-capturing classifier. This turns out to be an effective means of disassociating the target representation from the bias-capturing model and, as a result, from the protected attributes. 
To this end, the proposed method leverages the pairs of samples for which a typical task-specific loss is prone to bias, namely samples sharing either only targets or only protected attributes (see Figure \ref{fig:fig1}). This is not the case for previous BL-U methods that ignore the importance of a proper selection process, thereby impeding the bias mitigation effectiveness. Furthermore, it is theoretically justified that \METHOD\ can minimize the mutual information between the main model representation and protected attribute.

In summary, the main contributions of this paper are the following:
\begin{itemize}
    \item A fairness-aware representation learning approach that enables a model to learn fair representations w.r.t. a protected attribute by leveraging the representations of a bias-capturing classifier, without using the protected attribute labels.
    \item A condition for selecting the under-represented sample pairs that can contribute most effectively to the bias mitigation procedure.
    \item A wide comparative analysis involving 9 state-of-the-art approaches, 5 benchmark datasets, namely Biased-MNIST \cite{bahng2020rebias}, CelebA \cite{liu2015faceattributes}, UTK-Face \cite{zhifei2017cvpr}, Corrupted-Cifar10 \cite{hendrycks2018benchmarking} and 9-Class ImageNet \cite{bahng2020rebias}. In all the conducted experiments, \METHOD\ surpasses the current BL-U state-of-the-art by achieving +29.1\%, +18.1\%, +21.9\%, +4.2\%, and  +2.2\% accuracy on Biased-MNIST, CelebA, UTKFace, Corrupted-Cifar10, and 9-Class ImageNet, respectively. Moreover, in most experiments, \METHOD\ even outperforms the BL-A state-of-the-art.
\end{itemize}
The code is available at \url{https://github.com/gsarridis/FLAC}.
\section{Related Work}
\label{sec:related}
In recent years, many approaches have emerged for mitigating bias in neural networks. Some approaches focus on debiasing the training data by balancing the available data across the different groups of interest \cite{yan2020mitigating}, using more data sources \cite{gupta2018robot}, applying data augmentation \cite{liu2022reducing}, and collecting more data \cite{sattigeri2018fairness,xu2018fairgan}. However, such data-oriented approaches do not provide the flexibility required in cases where fairness requirements are not static (e.g., the need for considering a new protected attribute), while data availability may also be limited. 
To address these shortcomings, many efforts focus on developing approaches to mitigate bias while training a model using biased data (i.e., in-processing approaches \cite{ntoutsi2020bias}). To achieve this, many works propose ensembling \cite{clark2019LM,wang2020DI} or adversarial frameworks \cite{alvi2018turning,xie2017controllable,kim2019lnl,wang2019balanced,song2019learning,adel2019one}, contrastive learning-based approaches \cite{hong2021bb,barbano2022fairkl}, and regularization terms\cite{tartaglione2021end,cadene2019rubi,hong2021bb}. These in-processing bias mitigation approaches can be divided into two  primary categories that significantly impact their range of applicability and performance: methods that necessitate access to the protected attribute labels, i.e., BL-A, and methods that do not require such labels, i.e., BL-U.

\paragraph{Bias mitigation with protected attribute labels} 
In \cite{kim2019lnl} an adversarial approach is proposed, namely Learning Not to Learn (LNL), based on the MI between feature embeddings and protected attribute labels, that penalizes the model if it is capable of predicting the protected attributes. Domain-Independent (DI) \cite{wang2020DI} suggests multiple classifier heads, one per domain (i.e., bias type), for mitigating the bias in the feature space. The approach of Entangling and Disentangling deep representations (EnD) \cite{tartaglione2021end} proposes a regularization process that tries to entangle the feature vectors of samples with the same target class and disentangle the features representations of samples that share the same protected attribute label. 
Given the representations derived by the main model, FairKL \cite{barbano2022fairkl} aims to match the distances of positive (negative) bias-conflicting and bias-aligned samples from the anchor, where bias-conflicting and bias-aligned denote the under-represented and the over-represented groups, respectively.  When protected attribute labels are not available, the similarities between the biased features are utilized to assign weights to the FairKL regularizer. By contrast to FairKL, \METHOD\ solves a probability matching problem between the representations of a bias-capturing classifier and the main model.  Finally, the Bias-Contrastive and Bias-Balance (BC-BB) method \cite{hong2021bb} constitutes a contrastive learning-based approach for bias mitigation. Bias-Contrastive (BC) encourages the main model to pull the samples with the same target but different protected attributes closer in the feature space while ignoring the bias-aligned samples with different target labels. Bias-Balance (BB) can be used in combination with BC to further mitigate the bias by optimizing the model toward the data distribution. Although these methods demonstrate state-of-the-art performance in bias mitigation, they require the protected attribute labels to work. This constitutes a crucial limitation, as they can not be applied in many  real-world settings, where protected attribute labels are not available, or even worse, in cases where the bias is introduced by non-categorical visual attributes (e.g., texture).

\paragraph{Bias mitigation without protected attribute labels} The significant limitation described above, led the research community to explore methods that can address bias without being aware of the protected attribute labels. In \cite{clark2019LM}, the  Learned-Mixin (LM) approach is introduced, which proposes training the main model in an ensemble with the bias-capturing model in order to discourage the main model to encode the information that has been already captured by the bias-capturing model. Furthermore, Rubi \cite{cadene2019rubi} suggests a regularization term for adjusting the weighting of logits, thereby reducing the influence of biases during the training process. ReBias \cite{bahng2020rebias} is a framework that aims at the independence between the representations of the main model and the bias-capturing model using the Hilbert-Schmidt Independence Criterion \cite{gretton2005measuring}. The authors of \cite{nam2020LfF} suggest an approach, namely Learning from Failure (LfF), that uses the generalized Cross-Entropy (CE) loss for training the bias-capturing model to focus on simple samples that are likely to be aligned with the bias and in parallel, the training procedure followed for the main model encourages it to focus on the samples that the bias-capturing model fails to learn, which are expected to be bias-conflicting. 
The Spread Spurious Attribute (SSA) approach \cite{nam2022spread} leverages a limited set of bias-labeled samples to train a bias-capturing classifier. Subsequently, the predictions from this classifier are utilized to train a fair model through the minimization of the worst-group loss. In the same direction, the Confidence-based Group Label (CGL) \cite{jung2022learning} assignment methodology utilizes a protected attribute classifier to assign pseudo-protected attribute labels, while assigning random labels to low confidence samples. Then, CGL is combined with existing state-of-the-art approaches to improve their performance in terms of fairness.  An approach for diversifying the bias-conflicting samples through augmentation in the feature space (DistEnt) is proposed in \cite{lee2021learning}. Moreover, the BiasEnsemble (BE) \cite{lee2023revisiting} method proposes an unsupervised technique to discard the bias-conflicting samples from the training dataset. Empirical evidence indicates that this strategy can significantly improve the efficacy of multiple fairness-aware approaches.
Finally, the Soft-Contrastive (SoftCon) method \cite{hong2021bb}  constitutes an extension of the BiasCon \cite{hong2021bb}  that makes use of the similarities of the representations extracted by the bias-capturing model in order to enable BiasCon to be employed in scenarios where the protected attribute labels are not available. Although these methods successfully address the lack of protected attribute labels, they exhibit considerably lower performance compared to BL-A methods, as they do not consider appropriate mechanisms for focusing only on the samples that can effectively contribute to the bias mitigation, thus adding noise and preventing the model from learning fairer representations. 
In contrast, \METHOD\ is a BL-U method that minimizes the MI between the learned model and the protected attributes. This is achieved  through the use of a representation derived by a bias-capturing classifier (instead of the attribute labels) and a sampling strategy with optimality guarantees.


\paragraph{Mutual Information} MI has been used in various tasks, such as knowledge distillation  for preserving the teacher model's knowledge \cite{passalis2020heterogeneous,passalis2018learning,sarridis2022indistill}, feature selection  for selecting the features that are most related to the desired outcome \cite{vergara2014review}, clustering for assessing the clustering quality \cite{kraskov2005hierarchical}, and generative models for guiding generators in learning different manifolds \cite{li2021multi}. LnL \cite{kim2019lnl} proposed using MI in an adversarial framework for bias mitigation, but its performance is considerably lower than most of the bias mitigation approaches, due to the instability that adversarial approaches exhibit. Moreover, \cite{lum2016statistical} introduces a statistical framework that penalizes the MI between the target labels and the protected attribute labels.  An effort of converting the Quadratic Mutual Information (QMI) \cite{torkkola2003feature} problem to a probability matching problem for the knowledge distillation task is presented in \cite{passalis2018learning}. Inspired by this idea, we propose in this paper a method that leverages the capabilities of MI in order to effectively mitigate the bias, even in extreme data bias scenarios.

\section{Methodology}
In this section, we introduce the FLAC approach, which is based on correcting bias by focusing on sample representations where bias is identified. As illustrated in Figure \ref{fig:fig1}, these samples exhibit high (low) similarity for pairs with different (same) target labels and different (same) protected attribute labels, due to the attribute-class associations in the training data. A model demonstrating such behavior can function as a bias-capturing classifier, such as the Vanilla model or a model trained for protected attribute prediction. FLAC utilizes this insight, forcing the main model to showcase the converse behavior, specifically for these certain sample pairs (see Figure \ref{fig:fig1}). To this end, FLAC (i) suggests a sampling mechanism for the identification of relevant samples and (ii) solves a probability matching problem that encourages the main model to exhibit contrary behavior in terms of pairwise similarities to the bias-capturing model for the chosen sample pairs. Subsequent subsections detail the methodology and provide a theoretical analysis that justifies that minimizing the FLAC loss results in fair representations, that are independent of the protected attributes.
\subsection{Problem formulation}
The problem of mitigating bias in network representations can be formulated as follows. Let $(\mathbf{X}_i, y_i)$ be the $i$-th training sample of the dataset $\mathcal{D}$, where $\mathbf{X}_i$ is the input image, $y_i\in\mathcal{Y}$ the target, $h(\cdot)$ the model that we are interested to improve in terms of bias, and $b(\cdot)$ the bias-capturing model. Model $h(\cdot)$ is trained on the main task with targets $\mathcal{Y}$, while the bias-capturing model, $b(\cdot)$, is trained to predict protected attributes $t_i \in \mathcal{T}$ towards which the data (and consequently the model) is biased. Note that $t_i$ is not provided by $\mathcal{D}$, thus $b(\cdot)$ is trained on a different dataset (cf. Section~\ref{subsec:bias_capturing_model}). The representations extracted by the penultimate layer of the network are denoted as $\mathbf{h}_i$, while the corresponding bias-capturing classifier representations are denoted as $\mathbf{b}_i$. If the dataset $\mathcal{D}$ consists of samples that exhibit strong dependence between $\mathcal{Y}$ and $\mathcal{T}$, then training the model $h(\cdot)$ using only the task-specific loss function (e.g. CE for the multi-class classification task or a contrastive loss) will introduce bias into the model since it will rely on features that encapsulate information about $t_i$ for predicting $y_i$. Therefore, the goal is to eliminate the dependencies between representations $\mathbf{h}$  and $\mathbf{b}$.
\looseness=-1
\subsection{Bias-Capturing Classifier}\label{subsec:bias_capturing_model}
In many real-world scenarios, the attributes introducing bias are not categorical or the corresponding labels are not available. To address this issue, we make use of a bias-capturing classifier, as in \cite{bahng2020rebias,hong2021bb,barbano2022fairkl}. This acts as a feature extractor that encapsulates information related to the attributes that introduce bias to $h(\cdot)$.

If the protected attributes are categorical, the $b(\cdot)$ is trained on a dataset $\mathcal{D}'=\{(\mathbf{X}'_i,t_i)\}$ in which the protected attribute $t_i\in\mathcal{T}$ is provided (unlike $\mathcal{D}$). For instance, if \textit{gender} is the attribute related to bias, the bias-capturing classifier is trained to predict genders. In some cases, the model is biased towards attributes that are not categorical (e.g., texture or background) or there are no available datasets providing protected attribute annotations to train a bias-capturing model. Following previous literature \cite{bahng2020rebias,hong2021bb,barbano2022fairkl}, we opt for models trained on $\mathcal{D}$ that predict the task-specific targets $\mathcal{Y}$. For these networks, the penultimate layer is expected to adequately capture the attributes of interest \cite{bahng2020rebias}. To maintain clarity, when a Vanilla model is employed as a bias-capturing model, we label it as FLAC-B. Table \ref{tab:applic} outlines the applicability characteristics of FLAC variations in comparison to other BL-A and BL-U methods.
\begin{table}[ht]
    \centering
    \caption{Scenarios where FLAC variations can be applied compared to other BL-A and BL-U approaches.}
\begin{tabular}{lccc}
    \toprule
    \multirow{2}{*}{method} & \multicolumn{2}{c}{bias labels unavailable} & \multirow{2}{*}{non-categorical bias} \\ 
    \cline{2-3}
    & $\mathcal{D}$ & $\mathcal{D}'$ & \\ 
    \midrule
    BL-A & \xmark & n.a. & \xmark \\ 
    BL-U & \cmark & \cmark & \cmark \\ 
    \midrule
    FLAC & \cmark & \xmark & \xmark \\ 
    FLAC-B & \cmark & \cmark & \cmark \\ 
    \bottomrule
\end{tabular}
    
    \label{tab:applic}
\end{table}

\subsection{\METHOD}
\label{sec:method}
Discouraging the main model from learning to predict the protected attribute labels can be achieved by minimizing $\mathcal{I}(\mathbf{h}_i,t_i)$, where $\mathcal{I}(\cdot,\cdot)$ denotes MI. 
However, this is a task of high complexity and additionally it would require access to the protected attribute labels $\mathcal{T}$. 
Instead, we propose to use the pairwise similarity between the model representation and the representation derived by a bias-capturing classifier, i.e., a predictor of $t_i$, to minimize $\mathcal{I}(\mathbf{h}_i,t_i)$. More specifically, the target is to match the probability distributions of distances derived by the bias-capturing classifier features with the distributions of similarities derived by the representations of the main model for a certain subset of sample pairs. By doing so, we bring the samples with different protected attribute labels (i.e., $t_i\neq t_j$) and the same target label (i.e., $y_i=y_j$) closer to each other while increasing the distance between samples with the same protected attribute label (i.e., $t_i=t_j$) and different target labels (i.e., $y_i\neq y_j$).
For all possible pairs of batch samples' indices $(i,j)$ we define the following sets:
\begin{align*}
    \mathcal{A}_{10}&=\{(i,j)\mid y_i = y_j \land t_i \neq t_j\}\\
    \mathcal{A}_{01}&=\{(i,j)\mid y_i \neq y_j \land t_i = t_j\}\\
    \mathcal{A}_{11}&=\{(i,j)\mid y_i = y_j \land t_i = t_j,i\neq j\}\\
    \mathcal{A}_{00}&=\{(i,j)\mid y_i \neq y_j \land t_i \neq t_j\}\\
    \mathcal{A}&=\mathcal{A}_{10}\cup\mathcal{A}_{01}\cup\mathcal{A}_{11}\cup\mathcal{A}_{00}  
\end{align*} 
Then, involved pairs of samples should satisfy the following condition:
\begin{equation}
    (y_i = y_j \land t_i \neq t_j) \lor (y_i \neq y_j \land t_i = t_j)
    \label{eq:cond}
\end{equation}
forming the set $\mathcal{S}=\mathcal{A}_{10}\cup\mathcal{A}_{01}$.
 Note that involving all the possible sample pairs could have an adverse effect on the main model as it would lead to reducing (increasing) the similarity of samples with the same (different) target labels (see Section \ref{sec:analysis}). 
However, $\mathcal{T}$ labels are not available, thus a criterion needs to be defined for inferring whether two samples share the same protected attribute label. Let $K(\cdot)$ be a kernel function and $\mathcal{A}$ the set of all the possible pairs of samples, then the protected attribute equality is determined by:
\begin{equation}
    K(\mathbf{b}_i,\mathbf{b}_j)>  \frac{\max\limits_{u,v\in \mathcal{A}}{K(\mathbf{b}_u,\mathbf{b}_v)} + \min\limits_{u,v\in \mathcal{A}}{K(\mathbf{b}_u,\mathbf{b}_v)}}{2} \Rightarrow   
      t_i = t_j,~(i,j) \in \mathcal{A} 
\end{equation}
otherwise $t_i \neq t_j$. 
It should be noted that the proposed methodology will not be negatively impacted even if $K(\mathbf{b}_i,\mathbf{b}_j)$ falls below or exceeds this threshold for $(i,j)$ with $t_i=t_j$ or $t_i\neq t_j$, respectively. This is because the objective remains to enhance the similarity of $(i,j)$ where $y_i=y_j$ or reduce it when $y_i\neq y_j$.

\begin{figure}
    \centering
    \includegraphics[width=.61\linewidth,trim={0.5cm 0 0.2cm 0},clip]{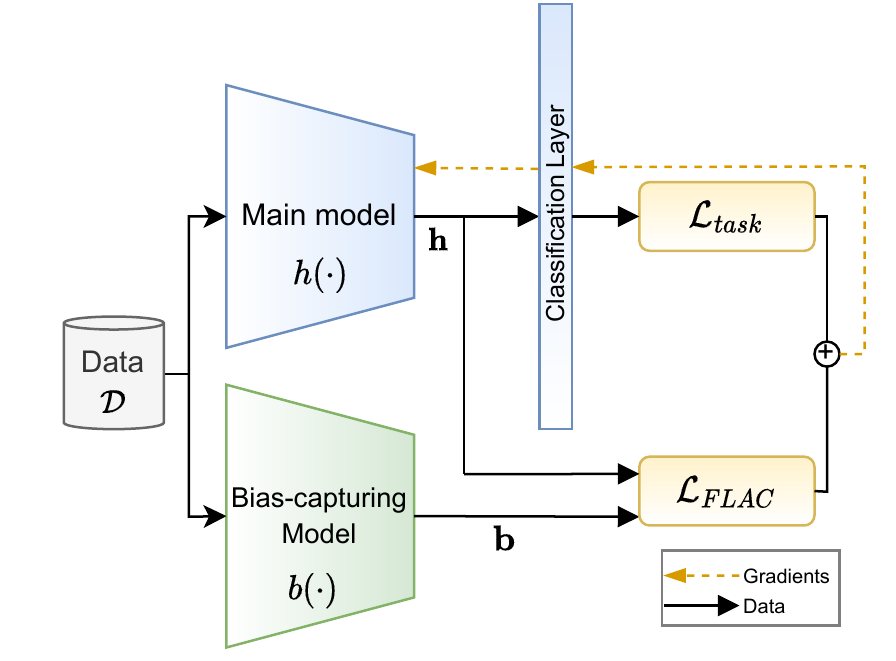}
    \caption[]{Illustration of the proposed framework.}
    \label{fig:frame}
\end{figure}

Then, the task-specific network's, $h(\cdot)$, probability distributions of the pairwise similarities can be defined as follows:
\begin{equation}
p_{i \mid j}^{(h)}=\frac{K(\mathbf{h}_i,\mathbf{h}_j)}{\sum_{k:(k,j)\in \mathcal{S}} K(\mathbf{h}_k,\mathbf{h}_j)} \in[0,1],~(i,j) \in \mathcal{S}.
\label{eq:dist_sim}
\end{equation}
Accordingly, for the bias-capturing model, we calculate the probability distributions of dissimilarities (i.e., ($1-K(\cdot)$):
\begin{equation}
p_{i \mid j}^{(b)}=\frac{1-K(\mathbf{b}_i,\mathbf{b}_j)}{\sum_{k:(k,j)\in \mathcal{S}} 1-K(\mathbf{b}_i,\mathbf{b}_j)} \in[0,1],~(i,j) \in \mathcal{S}.
\label{eq:dist_bcc}
\end{equation}
As regards the similarity kernel function, in this work we opted for the student's t kernel that demonstrates good performance on classification tasks \cite{passalis2020heterogeneous}:
\begin{equation}
    K(\mathbf{a},\mathbf{b}) = \frac{1}{1+||\mathbf{a} - \mathbf{b} ||_2}. 
\end{equation}
Having calculated the probability distributions of Equations \eqref{eq:dist_sim} and \eqref{eq:dist_bcc} for the pairs of samples in $\mathcal{S}$, our goal is to train a model that demonstrates high (low) similarity for the sample pairs that the bias-capturing classifier exhibit low (high) similarity. The divergence between these distributions can be calculated using Jeffreys divergence \cite{ali1966general}, a symmetric version of Kullback-Leibler divergence:
\begin{equation}
\mathcal{L}_{\METHOD}= 
\sum_{(i,j)\in\mathcal{S}} (p_{i \mid j}^{(b)} - p_{i \mid j}^{(h)}) \cdot 
 (\log p_{i \mid j}^{(b)} - \log p_{i \mid j}^{(h)}).
 \label{eq:jef}
\end{equation}
Then, the final loss can be defined as follows:
\begin{equation}\label{eq:loss}
    \mathcal{L} = \mathcal{L}_{task} + \alpha \cdot\mathcal{L}_{\METHOD},
\end{equation}
where $\alpha$ is a hyperparameter and $\mathcal{L}_{task}$ is the cost function of the task (e.g., the CE or a contrastive loss). Note that \METHOD\ fits better with pairwise losses as $\mathcal{L}_{task}$, due to its dependency on pairwise similarities. 
Figure \ref{fig:frame} presents the proposed framework.

\subsection{Theoretical Analysis}
\label{sec:analysis}
Let us denote with $\mathcal{B}=\{\mathbf{X}_i\}_{i=1}^N$ a batch with $N$ samples. 
Following the notation of Section \ref{sec:method}, 
we additionally consider the set $\Omega$ as a placeholder variable that upon declaration equals either $\mathcal{S}$ or $\mathcal{A}$. Finally, we assume that for a well-trained bias capturing classifier $\exists\epsilon>0$
, such that $\forall (i,j)\in\Omega$:
\begin{equation}
    t_i\neq t_j\Rightarrow K(\mathbf{b}_i,\mathbf{b}_j) \in [0,\epsilon)
    \label{eq:bias0}
\end{equation}
\begin{equation}
    t_i = t_j\Rightarrow K(\mathbf{b}_i,\mathbf{b}_j) \in (1-\epsilon,1]
    \label{eq:bias1}
\end{equation}

When the contrastive loss, denoted by $\mathcal{L}_{task}$, is minimized after training, then $\exists\delta_{10},\delta_{01},\delta_{11},\delta_{00},\epsilon'>0$ with $\epsilon'\le\delta_{10},\delta_{01},\delta_{11},\delta_{00}$, such that\footnote{
It has been shown that optimizing the standard cross-entropy loss: (i) corresponds to an approximate bound-optimizer of an underlying pairwise loss \cite{boudiaf2020unifying} and (ii) produces a feature space with inter-class distances being greater than intra-class ones \cite{das2019separability}. Nevertheless, accomplishing low $\delta_{10}$, $\delta_{01}$, $\delta_{11}$, $\delta_{00}$ applying to all $i,j$ is hardly possible. Hence, adopting the cross-entropy will only approximate the optimality derived by the contrastive loss.}: 
\begin{equation}\label{eq:k_set}
    K(\mathbf{h}_i,\mathbf{h}_j)\in \left\{
\begin{array}{ll}
      (1-\delta_{10}-\epsilon',1-\delta_{10}+\epsilon']\text{, if } (i,j)\in\mathcal{A}_{10} \\\relax
      [\delta_{01}-\epsilon', \delta_{01}+\epsilon')\text{, if } (i,j)\in\mathcal{A}_{01}\\
      (1-\delta_{11}-\epsilon',1-\delta_{11}+\epsilon']\text{, if } (i,j)\in\mathcal{A}_{11} \\\relax
      [\delta_{00}-\epsilon', \delta_{00}+\epsilon')\text{, if } (i,j)\in\mathcal{A}_{00} \\
\end{array} 
\right.
\end{equation}
 
However, in cases of existing bias in which the model has learned unwanted shortcuts (i.e., $\mathbf{h}_i\not\!\perp\!\!\!\perp t_i$), the magnitude of $\delta_{10},\delta_{01}$ monotonically depends on the amount of shared information between $\mathbf{h}_i$ and $t_i$\footnote{If the  shared information level is low, then there exist very small $\delta_{10},\delta_{01}$. If the  shared information level is high, then the $\delta_{10},\delta_{01}$ can indeed be high but the cardinality of $\mathcal{A}_{10}$ and $\mathcal{A}_{01}$ is very small, much smaller than the cardinality of $\mathcal{A}_{11}$ and $\mathcal{A}_{00}$ (cf. Section~\ref{sec:ablation} \& Figure~\ref{fig:num_of_pairs}).\label{fn:delta_1}}, while the magnitude of $\delta_{11},\delta_{00}$ not. This indicates the need to control $\delta_{10},\delta_{01}$ with an additional loss term that acts at least on $(i,j)\in\mathcal{S}$\footnote{An analysis showing that acting only on $\mathcal{S}$ is the best option in comparison to acting on the whole space $\mathcal{A}$, is considered below.}. To this end, in this work we propose the consideration of an additional loss term, namely $\mathcal{L}_{\METHOD}$, defined by Equation \eqref{eq:jef}, which will be minimized jointly with $\mathcal{L}_{task}$. This, then implies the following:

\begin{equation}
\begin{split}
    \mathcal{L}_{\METHOD}&=0 \Rightarrow \\
      \Rightarrow (p_{i \mid j}^{(b)} - p_{i \mid j}^{(h)}=0)&\lor(\log p_{i \mid j}^{(b)} - \log p_{i \mid j}^{(h)}=0) \Rightarrow \\
      \Rightarrow p_{i \mid j}^{(b)} &= p_{i \mid j}^{(h)} \Rightarrow \\
      \Rightarrow\frac{K(\mathbf{h}_i,\mathbf{h}_j)}{\sum_{k:(k,j)\in\Omega} K(\mathbf{h}_k,\mathbf{h}_j)}&=\frac{1-K(\mathbf{b}_i,\mathbf{b}_j)}{\sum_{k:(k,j)\in\Omega} 1-K(\mathbf{b}_k,\mathbf{b}_j)}  \Rightarrow \\
      \Rightarrow K(\mathbf{h}_i,\mathbf{h}_j)&=\alpha(j)\cdot (1-K(\mathbf{b}_i,\mathbf{b}_j))
\end{split}
\label{eq:K_alpha}
\end{equation}
where $(i,j)\in\Omega$. Considering for simplicity $\epsilon\approx\epsilon'\approx\delta_{11}\approx\delta_{00}\approx0$ and $\delta_{10}\approx\delta_{01}\overset{\text{def}}{=}\delta$, based on Equation \eqref{eq:k_set} we get:

\begin{equation}
    \begin{split}
        \alpha(j)&=\frac{\sum_{k:(k,j)\in\Omega} K(\mathbf{h}_k,\mathbf{h}_j)}{\sum_{k:(k,j)\in\Omega} 1-K(\mathbf{b}_k,\mathbf{b}_j)} =\frac{\sum_{k:(k,j)\in\mathcal{A}_{10}\cup\mathcal{A}_{01}\cup\mathcal{A}_{11}} K(\mathbf{h}_k,\mathbf{h}_j)}{\sum_{k:(k,j)\in\mathcal{A}_{10}\cup\mathcal{A}_{00}} 1-K(\mathbf{b}_k,\mathbf{b}_j)} =\\
        &=\frac{(1-\delta)\cdot\mid\mathcal{A}_{10}^j\mid+\delta\cdot\mid\mathcal{A}_{01}^j\mid+\mid\mathcal{A}_{11}^j\mid}{\mid\mathcal{A}_{10}^j\mid + \mid\mathcal{A}_{00}^j\mid} =\\ 
        &=\frac{\mid\mathcal{A}_{10}^j\mid + \mid\mathcal{A}_{11}^j\mid+\delta\cdot\big(\mid\mathcal{A}_{01}^j\mid-\mid\mathcal{A}_{10}^j\mid\big)}{\mid\mathcal{A}_{10}^j\mid + \mid\mathcal{A}_{00}^j\mid}
    \end{split}
    \label{eq:alpha}
\end{equation}
where $\mathcal{A}_{uv}^j=\{k\mid(k,j)\in\mathcal{A}_{uv}\}$ and $u,v\in\{0,1\}$.

Additionally, based on the Equations \eqref{eq:bias0}, \eqref{eq:bias1} and \eqref{eq:K_alpha}, minimizing Equation \eqref{eq:jef} results in:
\begin{equation}
    K(\mathbf{h}_i,\mathbf{h}_j) = \left\{
\begin{array}{ll}
      \alpha(j) & \text{, if } (i,j)\in\mathcal{A}_{10}\\
      0 &\text{, if } (i,j)\in\mathcal{A}_{01} \\
      0 & \text{, if } (i,j)\in\mathcal{A}_{11} \\
      \alpha(j) & \text{, if } (i,j)\in\mathcal{A}_{00}\\
\end{array} 
\right. 
\label{eq:k}
\end{equation}
So, if $\Omega=\mathcal{S}$, then $\mid\mathcal{A}_{11}^j\mid=\mid\mathcal{A}_{00}^j\mid=0$ implying $\alpha(j)=1$ (cf. Equation \eqref{eq:alpha})\footnote{Note that when bias is uniformly distributed across classes $\mid\mathcal{A}_{01}^j\mid$ = $\mid\mathcal{A}_{10}^j\mid$, which is the case in all experiments of this work. We consider exploring atypical cases of non-uniform bias distribution across classes as out of this works scope and leave it for future work.}, which results in an ideal form of Equation \eqref{eq:k}, namely:
\begin{equation}
    K(\mathbf{h}_i,\mathbf{h}_j) = \left\{
\begin{array}{ll}
      1 & \text{, if } (i,j)\in\mathcal{A}_{10}\\
      0 & \text{, if } (i,j)\in\mathcal{A}_{01} \\
\end{array} 
\right. 
\label{eq:k2}
\end{equation}
accomplishing the very small and independent $\delta_{10},\delta_{01}$ existence. In contrast, if $\Omega=\mathcal{A}$, then:
\[\mid\mathcal{A}_{11}^j\mid\ge 0\]
\[\mid\mathcal{A}_{00}^j\mid\ge 0\]
so depending on the sizes of the above sets $\alpha(j)$ can greatly vary. Moreover, not only the 3rd branch of Equation~\eqref{eq:k} forces pairs $(i,j)\in\mathcal{A}_{11}$ to have dissimilar embeddings $\mathbf{h}_i$ and $\mathbf{h}_j$ -thus lowering the accuracy-, but also the training becomes highly unstable with each batch imposing different $\alpha(j)$ (and consequently $K(\mathbf{h}_i,\mathbf{h}_j)$) values. 
Thus, we opt for considering $\mathcal{L}_{\METHOD}$ acting on $\Omega=\mathcal{S}$ to address bias and achieve higher accuracy levels.

By doing so, we argue that minimizing the total loss (i.e., Equation~\eqref{eq:loss}) minimizes mutual information between the main model's representations $\mathbf{h}$ and the corresponding protected attributes $t$. In order to prove this, we consider the definition of QMI as defined in \cite{torkkola2003feature}, namely Equation \eqref{eq:qmi_int}: 
\begin{equation}
\begin{split}
    \mathcal{I}(\mathbf{h}, t)&=\sum_t \int_{\mathbf{h}} p(\mathbf{h}, t)^2 d\mathbf{h}+\sum_t \int_{\mathbf{h}}(p(\mathbf{h}) P(t))^2 d \mathbf{h}- \sum_t \int_{\mathbf{h}} p(\mathbf{h}, t) p(\mathbf{h}) P(t) d \mathbf{h},
\end{split}    
\label{eq:qmi_int}
\end{equation}
whose 3 terms are called information potentials and are denoted by $V_{IN}$, $V_{ALL}$, and $V_{BTW}$, respectively.  
Then, the information potentials can be estimated as follows:
\begin{equation}
    V_{IN} = \frac{1}{N^2}\sum_{p=1}^{N_c} \sum_{k=1}^{J_p} \sum_{l=1}^{J_p} K(\mathbf{h}_{pk}, \mathbf{h}_{pl}),
    \label{vin}
\end{equation}
\begin{equation}
    V_{ALL} = \frac{1}{N^2}\sum_{p=1}^{N_c} (\frac{J_p}{N})^2 \sum_{k=1}^{N} \sum_{l=1}^{N} K(\mathbf{h}_{k}, \mathbf{h}_{l}),
    \label{vall}
\end{equation}
\begin{equation}
    V_{BTW} = \frac{1}{N^2}\sum_{p=1}^{N_c} \frac{J_p}{N} \sum_{j=1}^{J_p} \sum_{k=1}^{N} K(\mathbf{h}_{pj}, \mathbf{h}_{k}),
    \label{vbtw}
\end{equation}
where $N$, $N_c$, and $J_p$ are the number of samples, classes, and samples belonging to the class $p$, respectively. Note that $J_p$ is equal to $\frac{N}{N_c}$ as QMI should be calculated on balanced data.  $V_{IN}$ consists of the interactions between pairs within each protected attribute class, $V_{ALL}$ consists of the interactions between all pairs, and $V_{BTW}$ consists of interactions between samples of each protected attribute class against all samples. Considering the optimal scenario with $\mathcal{L}_{task}=0$ and $\mathcal{L}_{\METHOD}=0$, the Equation \eqref{eq:k_set} holds for very small $\delta_{10}$, $\delta_{10}$, $\delta_{11}$, $\delta_{00}$ (for simplicity we consider $\delta_{10}\approx\delta_{10}\approx\delta_{11}\approx\delta_{00}\approx\epsilon'\approx0$) and the information potentials, $V_{IN}$, $V_{ALL}$, and $V_{BTW}$, result in the following: 
\begin{equation}
    V_{IN} = \frac{1}{N^2} \mid\mathcal{A}_{11}\mid,
\end{equation}
\begin{equation}
\begin{split}
    V_{ALL} & = \frac{1}{N^2} \sum_{p=1}^{N_c} \left(\frac{J_p}{N}\right)^2 \left( \mid\mathcal{A}_{11}\mid + \mid\mathcal{A}_{10}\mid \right) =  \overset{J_p=\frac{N}{N_c}} =  \frac{1}{N^2} \frac{1}{N_c} \left( \mid\mathcal{A}_{11}\mid + \mid\mathcal{A}_{10}\mid \right),
\end{split}
\end{equation}
\begin{equation}
\begin{split}
    V_{BTW} & = \frac{1}{N^2} \sum_{p=1}^{N_c} \frac{J_p}{N} \frac{\mid\mathcal{A}_{11}\mid + \mid\mathcal{A}_{10}\mid}{N_c} =  \overset{J_p=\frac{N}{N_c}} =  \frac{1}{N^2} \frac{1}{N_c} (\mid\mathcal{A}_{11}\mid + \mid\mathcal{A}_{10}\mid).
\end{split}
\end{equation}
Finally, given that $\mid\mathcal{A}_{10}\mid = (N_c-1)\cdot\mid\mathcal{A}_{11}\mid$, in the optimal scenario of $\mathcal{L}=0$ the QMI between the representations $\mathbf{h}$ and the corresponding protected attributes $t$, is calculated as:
\begin{equation}
    \begin{split}
         \mathcal{I}(\mathbf{h}, t) & = V_{IN} + V_{ALL} - 2 V_{BTW} = V_{IN} -  V_{BTW} =\\
        & = \frac{1}{N^2} \left(\mid\ \mathcal{A}_{11} \mid - \frac{\mid\ \mathcal{A}_{11} \mid + \mid\ \mathcal{A}_{10} \mid}{N_c} \right) =\\
        & = \frac{1}{N^2}  \left(\mid\ \mathcal{A}_{11} \mid  - \frac{\mid\ \mathcal{A}_{11} \mid  + (N_c-1)\mid\ \mathcal{A}_{11} \mid }{N_c} \right)= \\
        & =\frac{1}{N^2}  \left(\mid\ \mathcal{A}_{11} \mid  -  \frac{\mid\ \mathcal{A}_{11} \mid }{N_c} -  \frac{(N_c-1)\mid\ \mathcal{A}_{11} \mid }{N_c}\right) =\\
        & = \frac{1}{N^2} \left( \left(1-\frac{1}{N_c}\right) \mid\ \mathcal{A}_{11} \mid  -  \frac{(N_c-1)}{N_c} \mid\ \mathcal{A}_{11} \mid  \right) =\\
        & = 0 
    \end{split}
\end{equation}
Thus, \METHOD\ can indeed lead to fair representations, that are independent of the protected attributes.
\section{Experimental setup}
\subsection{Datasets}
\paragraph{Datasets with Artificially Injected Bias} Biased-MNIST constitutes a biased version of the original MNIST dataset \cite{lecun1998mnist}, introduced in \cite{bahng2020rebias} as a standard benchmark for evaluating the effectiveness of bias mitigation methods. It consists of digits with colored backgrounds (10 colors in total). The bias is introduced through the association of each digit with a certain color. The probability of a digit having biased background is denoted as $q$, while a random background color is assigned with probability $1-q$. This way, higher $q$ values lead to a stronger association between digits and colors and thus more biased data. Following previous works using this dataset, here we consider four variations of the Biased-MNIST w.r.t. values of $q$, namely 0.99, 0.995, 0.997, and 0.999.

\paragraph{Established computer vision datasets} We also evaluate the proposed approach on three established computer vision datasets, namely CelebA \cite{liu2015faceattributes}, UTK-Face \cite{zhifei2017cvpr}, and 9-Class ImageNet \cite{bahng2020rebias}. As regards the CelebA dataset that consists of more than 200,000 facial images annotated with 40 binary attributes, the \textit{gender} is the protected attribute, while the \textit{HeavyMakeup} and the \textit{BlondHair} constitute the two target labels, as in \cite{hong2021bb}. For the UTKFace, which is a dataset consisting of over 20,000 facial images with \textit{gender}, \textit{race}, and \textit{age} labels, we consider the \textit{gender} as the target label, while \textit{race} and  \textit{age} as the protected attributes. For UTKFace, we enforce a dependence of 90\% between the target and the protected attribute  following the experimental setup of \cite{hong2021bb}.
The Corrupted-Cifar10 \cite{hendrycks2018benchmarking} dataset consists of 10 classes with texture biases evident in the training data and it offers four correlation ratios: 0.95, 0.98, 0.99, and 0.995.
The 9-Class ImageNet, a subset of the ImageNet dataset \cite{deng2009imagenet} consisting of 9 super-classes, is employed to evaluate the proposed method in settings where bias is not explicitly associated with a given attribute (e.g., texture bias). In addition, the ImageNet-A \cite{hendrycks2021nae}, which comprises ImageNet samples that Vanilla models often fail to classify, is also involved in the experiments as a test set. 

\subsection{Model architectures}
\label{sec:models}
For the controlled experiments conducted for Biased-MNIST, we employ the CNN architecture proposed in \cite{bahng2020rebias}, namely \textit{SimpleConvNet} that consists of four convolutional layers with 7$\times$7 kernels and a fully connected layer with 128 neurons. For the CelebA, UTKFace, Corrupted-Cifar10, and 9-Class ImageNet, we employ the Resnet-18 \cite{he2016deep} architecture for both the main and the bias-capturing models, except for the 9-Class ImageNet, where we use the BagNet-18 \cite{brendel2019approximating} as the bias-capturing model following the practice of \cite{bahng2020rebias,hong2021bb,barbano2022fairkl}. BagNets mainly consist of 1$\times$1 convolutions  and they are found to be more prone to texture bias \cite{bahng2020rebias}. Note that all model selections are based on the previous literature to ensure comparability.

\subsection{Baseline methods}
The proposed method is compared to 9 state-of-the-art approaches. Of those, BL-U approaches include the LM \cite{clark2019LM}, Rubi \cite{cadene2019rubi}, ReBias \cite{bahng2020rebias}, LfF \cite{nam2020LfF}, and SoftCon \cite{hong2021bb}. The BL-A approaches considered in our comparative study are the LNL \cite{kim2019lnl}, EnD \cite{tartaglione2021end}, DistEnt \cite{lee2021learning}, BE \cite{lee2023revisiting}, BC-BB \cite{hong2021bb}, and FairKL \cite{barbano2022fairkl}. More details about the competing methods can be found in Section \ref{sec:related}. For ensuring a fair comparison between the approaches, it should be highlighted that: 
\begin{itemize}
    \item \METHOD\ combined with a Vanilla biased model as a bias-capturing model (i.e., FLAC-B) allows for a direct comparison with BL-U approaches, devoid of any reliance on protected attribute labels.
    \item  Conversely, \METHOD\ combined with a bias-capturing model trained using attribute labels of $\mathcal{D}'$ holds an advantage over BL-U methods that do not use any extra data and a disadvantage compared to BL-A methods due to the error of the bias-capturing models in predicting the protected attributes.
\end{itemize}


\subsection{Implementation details and evaluation protocol}
We employ the Adam optimizer for all the experiments with an initial learning rate of 0.001 that decays by a factor of 0.1 at 1/3 and 2/3 of the total training epochs, the weight decay is equal to $10^{-4}$ and the batch size is set to 128, unless stated otherwise. 
For the Biased-MNIST, models are trained for 80 epochs. The values of $\alpha$ for each BiasedMNIST variant, 0.99, 0.995, 0.997, and 0.999 are 110, 1500, 2000, and 10,000 respectively. Note that such large values of $\alpha$ were selected due to the large discrepancy between the values of $\mathcal{L}_{\METHOD}$ and $\mathcal{L}_{task}$. The $\alpha$ values were chosen following an initial grid search. There is a possibility that a more comprehensive hyperparameter tuning process could lead to improved outcomes. Additionally, it has been noticed that small $\alpha$ can lead to unstable training behavior. The augmentations applied for the established computer vision datasets (i.e., CelebA, UTKFace, and 9-Class ImageNet) are the \textit{random resized crop} and \textit{random horizontal flip}. For the CelebA, models are trained for 40 epochs with $\alpha$ equal to 20,000 for the \textit{HeavyMakeup} classification task, while 10 training epochs with $\alpha$ equal to 30,000 are used for the \textit{BlondHair} classification task. The images are resized to 224$\times$224. For UTKFace, models are trained for 20 epochs, $\alpha$ is set to 1,000 and image size to 64$\times$64. For the experiments leveraging bias-capturing models trained with the protected attributes as targets, we utilized the FairFace dataset which provides \textit{gender}, \textit{race}, and \textit{age} annotations. Specifically, the models for \textit{race}, \textit{age}, and \textit{gender} demonstrate 89.86\%, 98.80\%, and 95.79\% accuracy on UTKFace and CelebA, respectively. For Corrupted-Cifar10, the models are trained for 100 epochs using a cosine annealing scheduler, $\alpha$ is equal to 100,000, and image size is 32$\times$32. For 9-Class ImageNet, the cosine annealing learning rate scheduler is used, the number of training epochs is equal to 250, $\alpha$ is set to 1,000, and image size to 224$\times$224. As $\mathcal{L}_{task}$ we consider both the Supervised Contrastive (SupCon) \cite{khosla2020supervised} and the CE loss across all the experiments in order to provide a fair comparison between the competing bias mitigating methods. In addition, only subtle performance discrepancies have been noticed when assigning a weight to this loss. All the experiments were conducted on a single NVIDIA RTX-3090 GPU and repeated for 5 different random seeds (all results reported in Section \ref{sec:results} refer to the corresponding mean scores).

The test set used for BiasedMNIST evaluation has $q=0.1$ in order to be unbiased (i.e., no association between digits and colors). For CelebA and UTKFace, we use the unbiased and bias-conflict test sets provided by \cite{hong2021bb} for the evaluation. For Corrupted-Cifar10, the official unbiased test set is considered. Finally, for the 9-Class ImageNet experiments, two test sets are involved, (a) the official 9-Class ImageNet test set and (b) the ImageNet-A. For all the above, accuracy is employed as an evaluation metric. In addition, we measure the unbiased accuracy on 9-Class ImageNet test set by using the texture bias annotations provided by \cite{bahng2020rebias} and averaging the accuracy calculated for each one of the texture groups. Furthermore, apart from the typical performance metric (i.e, accuracy) we also report the performance of \METHOD\ in terms of three fairness metrics for disparate impact and disparate mistreatment, namely \textit{p\% rule} \cite{meier1984happened}, Difference in False Positive Rates (DFPR), and Difference in False Negative Rates (DFNR) \cite{janssen2021bias,krasanakis2018adaptive}.

\section{Results}
\label{sec:results}
\subsection{Controlled experiments}
Table \ref{tab:mnist} presents the performance comparison of the proposed approach against 5 BL-U and 4 BL-A methods. \textit{Vanilla} refers to the model performance using only the CE loss without considering any bias mitigation algorithm. According to Table \ref{tab:mnist}, the proposed method consistently outperforms all competing methods, even the BL-A ones. In particular, \METHOD\ enhances the accuracy by 0.1\%-0.6\% for the different $q$ values compared to the best performing BL-A method and by 3.5\%-29.1\% compared to the best performing BL-U approach. Furthermore, it is worth noting that while $q$ increases, \METHOD\ is capable of maintaining very high accuracy, which is not the case for most of the compared methods. For instance, for the extreme $q$ value of 0.999, \METHOD\ achieves 94.1\% accuracy, while the best-performing BL-U method achieves 65\%. In addition, note that \METHOD\ combined with CE loss significantly outperforms the corresponding competing methods that use CE as $\mathcal{L}_{task}$. As regards the selection of $\mathcal{L}_{task}$, as expected opting for a pairwise loss (i.e., SupCon) results in improved accuracy as justified in Section \ref{sec:analysis}. 

\begin{figure}[H]
        \centering
        \begin{subfigure}[t]{0.49\linewidth}
            \centering
            \includegraphics[width=0.9\linewidth,trim={0.3cm 0.6cm 1.2cm 1.1cm},clip]{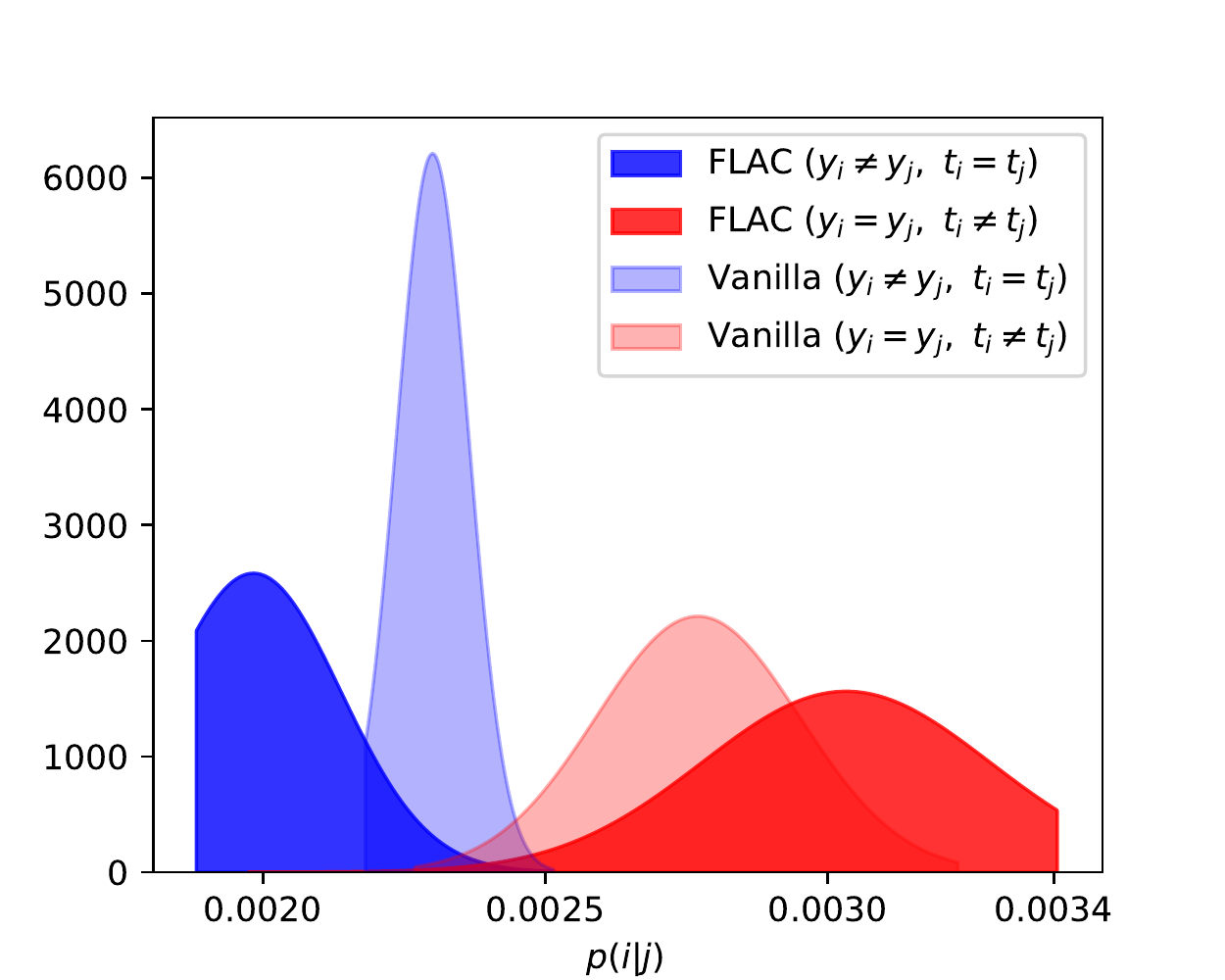}
            \caption[]%
            { \scriptsize{Distributions of Equation \eqref{eq:dist_sim} for Vanilla and \METHOD\ trained on Biased-MNIST with $q=0.99$ and achieves 88.9\% and 96.9\% accuracy, respectively. Using \METHOD, the similarity between samples with the same or different label is increased or decreased respectively, compared to Vanilla.}}  
        \end{subfigure}
        \begin{subfigure}[t]{0.49\linewidth}   
            \centering 
            \includegraphics[width=.9\linewidth,trim={0.3cm 0.6cm 1.2cm 1.1cm},clip]{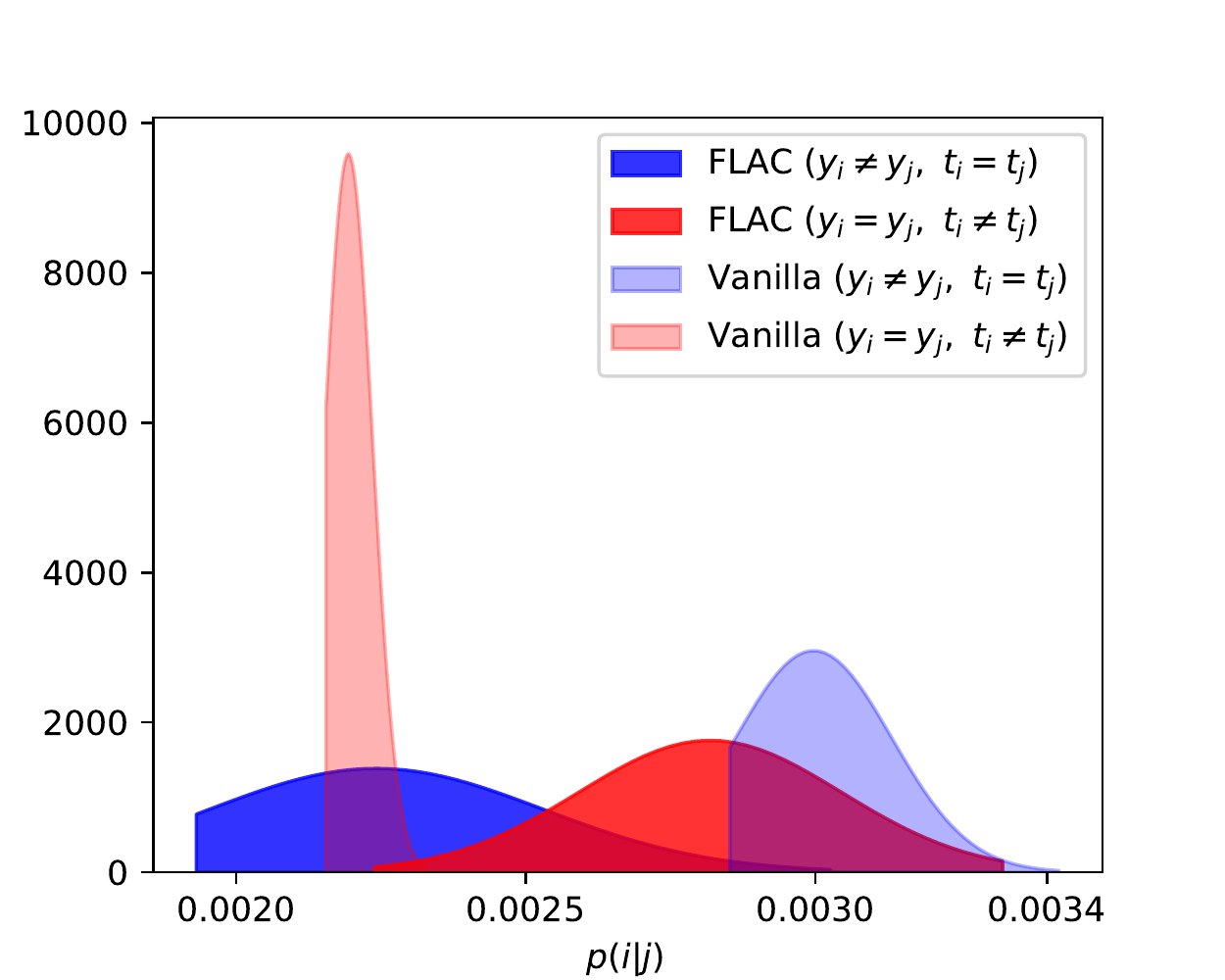}
            \caption[]%
            {\scriptsize{Distributions of Equation \eqref{eq:dist_sim} for Vanilla (11.8\% acc.) and \METHOD\ (89.3\% acc.) trained on Biased-MNIST with $q=0.999$. In this evaluation setup, Vanilla results in an extremely biased model that exhibits high (low) similarities for samples with the same (different) protected attribute label and different (same) target. Here, the \METHOD's impact is significant as it manages to flip the distributions.  }    }
        \end{subfigure}

        \caption[ ]{The distributions of similarities or/and distances for the samples that satisfy Equation \eqref{eq:cond}. \textit{Red} color represents the sample pairs with the same target, but different protected attribute label, while \textit{blue} depicts the pairs with different target, but the same protected attribute labels.} 
        \label{fig:dist}
    \end{figure}

As shown in Figure \ref{fig:dist} (a), which presents the distributions (with $q=0.99$) derived by Equation \eqref{eq:dist_sim} for Vanilla and \METHOD\, the proposed method increases (decreases) the similarities of samples with the same (different) target labels and different (same) protected attribute labels compared to Vanilla. Furthermore, for $q=0.999$, where Vanilla results in a model that learned the background colors instead of the digits, \METHOD\ is capable of learning the proper distributions as depicted in Figure \ref{fig:dist} (b).

\begin{table}[h]
\begin{center}
\caption{Evaluation on Biased-MNIST for different bias levels. 
Underlined and dotted-underlined values refer to the second best performing BL-U and the best performing BL-A methods, respectively. Con. refer to Contrastive losses. * refers to the performance with color-jittering (i.e., randomly changing the brightness, contrast, saturation, and hue of images). }\label{tab:mnist}
      \resizebox{0.78\linewidth}{!}{
    \begin{tabular}{lcccccc} 
          \toprule
    \multirow{ 2}{*}{methods} & \multirow{ 2}{*}{$L_{task}$}  & \multirow{ 2}{*}{BL-U} & \multicolumn{4}{c}{$q$} \\ 
    \cline{4-7}
 & & & 0.99 & 0.995 & 0.997 & 0.999 \\ \midrule
 Vanilla & CE & \cmark & 90.8$\pm$0.3 & 79.5$\pm$0.1 & 62.5$\pm2.9$ & 11.8$\pm$0.7 \\ 
         LNL& CE & \xmark& 86.0$\pm$0.2 & 72.5$\pm$0.9 & 57.2$\pm$2.2 & 18.2$\pm$1.2\\
         EnD& CE & \xmark&  94.8$\pm$0.3 & 94.0$\pm$0.6 &  82.7$\pm$0.3 & 59.5$\pm$2.3\\
         
         BC-BB & Con.* & \xmark& \dotuline{98.1}$\pm$0.1& \dotuline{97.7}$\pm$0.1 &  \dotuline{97.3}$\pm$0.1 & \dotuline{94.0}$\pm$0.6 \\
         FairKL & CE & \xmark& 95.9$\pm$0.2 & 94.8$\pm$0.5 & 93.9$\pm$1.1 & 79.9$\pm$4.3\\ 
         FairKL & Con. & \xmark& 97.9$\pm$0.0 & 97.0$\pm$0.0 & 96.2$\pm$0.2 & 90.5$\pm$1.5\\ 
         LM & CE &\cmark& 91.5$\pm$0.4 & 80.9$\pm$0.9 & 56.0$\pm$4.3 & 10.5$\pm$0.6\\
         Rubi & CE &\cmark&  85.9$\pm$0.1 & 71.8$\pm$0.5 & 49.6$\pm$1.5 & 10.6$\pm$0.5\\
         ReBias & CE &\cmark& 88.4$\pm$0.6 & 75.4$\pm$1.0 & 65.8$\pm$0.3 & 26.5$\pm$1.4\\
         LfF & CE &\cmark& 95.1$\pm$0.1 & 90.3$\pm$1.4 & 63.7$\pm$20.3 & 15.3$\pm$2.9\\
         DisEnt\textsuperscript{\ref{reimp}} & CE & \cmark& 96.1$\pm$0.4 & 91.5$\pm$0.3 & 81.9$\pm$6.3 & 38.3$\pm$20.1 \\
         DisEnt + BE\textsuperscript{\ref{reimp}} & CE &\cmark& 94.0$\pm$1.7 & 90.8$\pm$3.0 & 80.9$\pm$11.2 & 26.4$\pm$2.4\\
         SoftCon & Con. &\cmark& 95.2$\pm$0.4 & 93.1$\pm$0.2 & 88.6$\pm$1.0 & 65.0$\pm$3.2 \\ \midrule
         \METHOD & CE &\cmark& 96.9$\pm$0.0  & {94.9}$\pm$0.5 & {94.1}$\pm$0.5 & {89.3}$\pm$1.3\\ 
         \METHOD & Con. &\cmark& \underline{97.9}$\pm$0.1  & \underline{96.8}$\pm$0.0 & \underline{95.8}$\pm$0.2 & \underline{89.4}$\pm$0.8\\
         \METHOD & Con.* &\cmark& \textbf{98.7}$\pm$0.0  & \textbf{98.2}$\pm$0.1 & \textbf{97.8}$\pm$0.1 & \textbf{94.1}$\pm$1.1 \\
         \bottomrule
         
\end{tabular}
  }
\end{center}
\end{table}

\subsection{Evaluation on established datasets}
Table \ref{tab:celeba} compares the methods for the two tasks of CelebA, namely predicting \textit{BlondHair} and \textit{HeavyMakeup}. The \textit{unbiased} and the \textit{bias-conflict} refer to test sets that have balance bias-aligned and bias-conflict samples and only bias-conflict samples, respectively. As regards the \textit{BlondHair} prediction task, the proposed method outperforms the best competing BL-U method by 7\% and 7.5\% on the \textit{unbiased} and the \textit{bias-conflict} test sets, respectively. In addition, \METHOD\ manages to outperform even the best competing BL-A method by 1.5\% on the \textit{bias-conflict} test set, while demonstrating competitive performance on the \textit{unbiased} test set (i.e., -0.2\%). It should be stressed that FLAC attains this competitive performance despite the bias-capturing model's errors, unlike BL-A methods which directly utilize bias labels. As regards the most challenging task, namely \textit{HeavyMakeup}, the proposed method significantly outperforms both the BL-A and the BL-U approaches. In particular, \METHOD\ achieves 85.4\% and 79.1\% accuracy on  the \textit{unbiased} and the \textit{bias-conflict} test sets, while the best performing BL-A method only manages to get 78.6\% and 63.5\% in terms of accuracy, respectively. Here, \METHOD\ with CE demonstrates higher performance compared to \METHOD\ combined with SupCon loss. Finally, the FLAC's performance when employing the Vanilla model as a bias capturing classifier is reported as FLAC-B in Table \ref{tab:celeba}. It can be easily noticed that FLAC-B demonstrates significantly improved performance compared to the rest of BL-U approaches while being completely blind to the protected attribute labels. Moreover, for the \textit{HeavyMakeup} task, FLAC-B notably outperforms the BL-A methods across both \textit{unbiased} and \textit{bias-conflict} samples. 

\begin{table}[t]
\begin{center}
 \caption{Evaluation of the proposed method on CelebA for two different target attributes, namely \textit{HeavyMakeup} and \textit{BlondHair}, with \textit{gender} as the protected attribute.}
    \label{tab:celeba}
\begin{tabular}{lcccccc} 
\toprule
         \multirow{ 3}{*}{methods} & \multirow{ 3}{*}{$L_{task}$}  &  \multirow{ 3}{*}{BL-U} & \multicolumn{4}{c}{target}   \\ \cline{4-7}
       &  & & \multicolumn{2}{c}{BlondHair} &  \multicolumn{2}{c}{HeavyMakeup} \\ \cline{4-7}
        & & & unbiased & bias-conflict & unbiased & bias-conflict \\ \midrule
Vanilla & CE & \cmark & 79.0$\pm$0.1 & 59.0$\pm$0.1 & 76.0$\pm$0.8 & 55.2$\pm$1.9 \\ 
         LNL & CE & \xmark& 80.1$\pm$0.8  & 61.2$\pm$1.5 & 76.4$\pm$2.3  & 57.2$\pm$4.6 \\ 
         
         DI & CE & \xmark& 90.9$\pm$0.3 & 86.3$\pm$0.4 & 74.3$\pm$1.1  & 53.8$\pm$1.6 \\ 
         EnD & CE & \xmark& 86.9$\pm$1.0 & 76.4$\pm$1.9 & 74.8$\pm$1.8 & 53.3$\pm$3.6 \\ 
         BC-BB  & Con. & \xmark& \dotuline{91.4}$\pm$0.0 & \dotuline{87.2}$\pm$0.2 & \dotuline{78.6}$\pm$1.8 & \dotuline{63.5}$\pm$3.7 \\ 
         FairKL\footnotemark & Con. & \xmark & 81.7$\pm$1.7 & 69.9$\pm$2.4 & 77.4$\pm$1.1  & 57.2$\pm$1.6 \\
         LfF & CE &\cmark & 84.2$\pm$0.3 &  81.2$\pm$1.4 & 66.2$\pm$1.2 &  45.5$\pm$4.3\\ 
         SoftCon & Con. &\cmark & 84.1 & 74.4 & 77.4 & 61.0 \\ \midrule
         \METHOD & CE &\cmark &  \underline{90.1}$\pm$0.3 &  \underline{87.6}$\pm$0.6 & \textbf{85.4}$\pm$1.9 & \underline{79.1}$\pm$4.1 \\
         \METHOD & Con. &\cmark & \textbf{91.2}$\pm$0.3& \textbf{88.7}$\pm$0.5 & \underline{84.7}$\pm$1.7 & 78.8$\pm$4.6 \\
         \METHOD-B & Con. &\cmark & 87.0$\pm$0.6 & 84.9$\pm$2.2 & 82.6$\pm$3.0 & \textbf{79.4}$\pm$4.1 \\
\bottomrule
\end{tabular} 
\end{center}
\end{table}
\footnotetext{Reimplemented based on the code provided by the authors. \label{reimp}}
Table \ref{tab:utk} presents the results of \METHOD\ compared to other methods on the UTKFace dataset with \textit{race} and \textit{age} as protected attributes. For both protected attributes, the proposed method surpasses the best performing competing methods on both the \textit{unbiased}  (i.e., +1\% and +1.6\%) and \textit{bias-conflict} test sets (i.e., 4\% and 9.4\%), while it outperforms the state-of-the-art BL-U methods by 5\%-6.1\% and  13\%-21.9\% on the \textit{unbiased} and \textit{bias-conflict} test sets, respectively. Here, it is noteworthy that \METHOD\ exhibits very high accuracy (i.e., 81.1\%) on the most challenging test set (i.e., \textit{bias-conflict} test set with \textit{age} as the protected attribute), whereas the best BL-A method achieves 71.7\%. In addition, FLAC-B not only maintains high unbiased accuracy but also achieves notable improvements in the performance on bias-conflicting samples, even though it employs the Vanilla model as a bias-capturing model.

\begin{figure*}[t]
        \centering

        \begin{subfigure}[t]{0.4\linewidth}
            \centering
            \includegraphics[width=\linewidth,trim={2cm 2cm 2cm 2cm},clip]{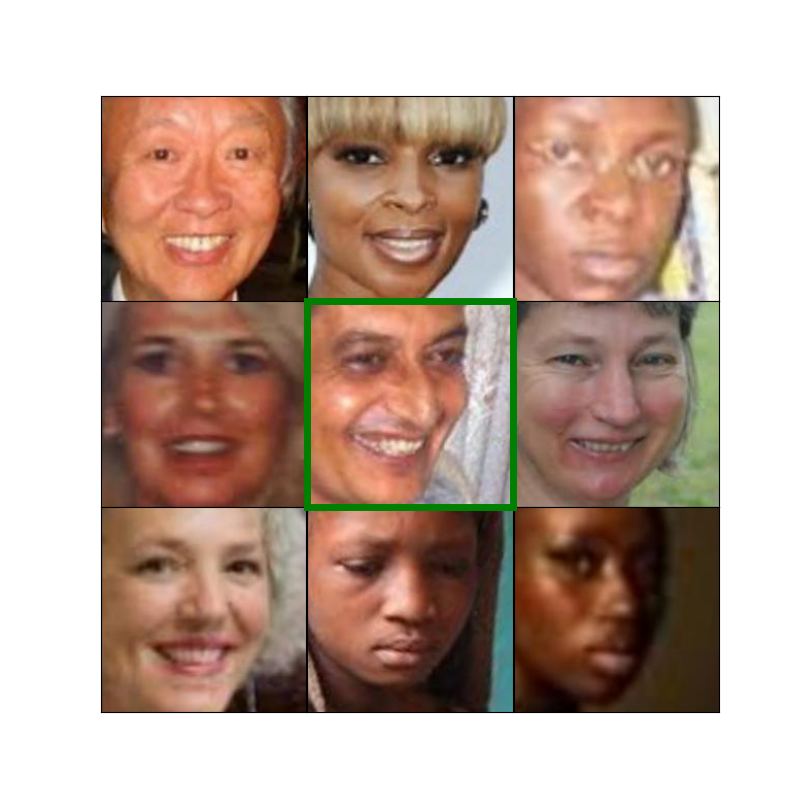}
            \caption[]{Method: Vanilla, Query image: non-white male} 
            \label{fig:vis_van_01}
        \end{subfigure}
        ~
        \begin{subfigure}[t]{0.4\linewidth}   
            \centering 
            \includegraphics[width=\linewidth,trim={2cm 2cm 2cm 2cm},clip]{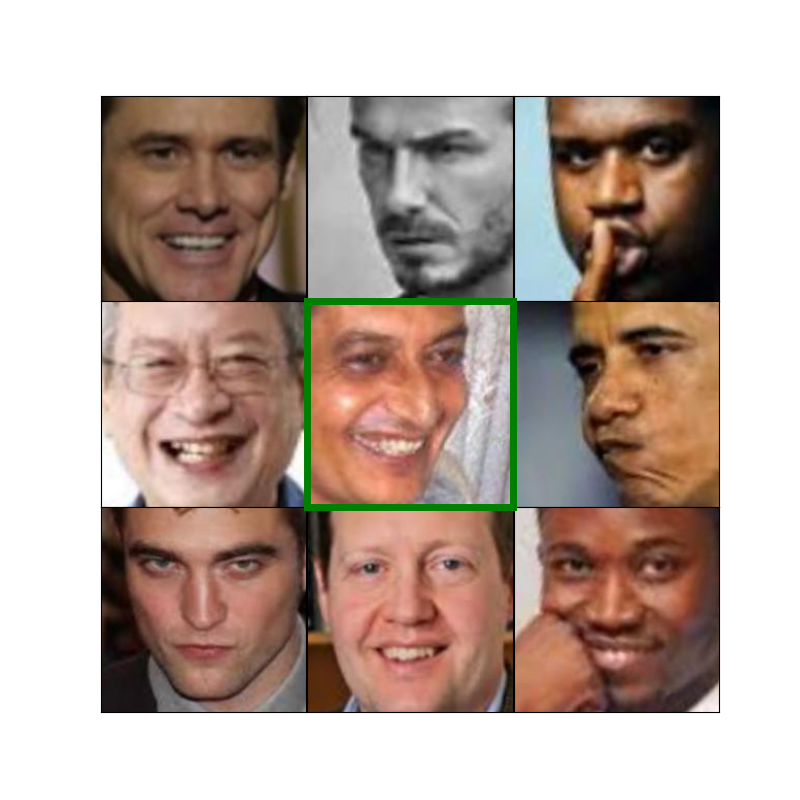}
            \caption[]{Method: \METHOD, Query image: non-white male}  
            \label{fig:vis_fluo_01}
        \end{subfigure}        
        

        \caption[ ]{The top 8 images that are most similar to queries representing a minority group of UTKFace (i.e., non-white males) based on the representations derived by Vanilla and \METHOD\ approaches. Images with green borders denote the query samples.}
        \label{fig:vis}
    \end{figure*}

Furthermore, Figure \ref{fig:vis} visualizes the 8 most similar samples to a query image that represents a minority group for Vanilla and \METHOD. In this example, we employed the biased UTKFace dataset with \textit{gender} as target and \textit{race} as protected attribute, where the minority group is the non-white males, due to the high correlation between the males and the white race in the training data.  By inspecting Figures \ref{fig:vis_van_01} and \ref{fig:vis_fluo_01}, we may observe that given a query depicting a \textit{non-white male}, Vanilla's similar representations correspond mostly to people of color (both males and females) while retrieved images based on the \METHOD's representations depict males of various races, i.e., they do not encode racial information.
The visualization provided in Figure \ref{fig:vis} indicates the effectiveness of \METHOD\ in (i) disassociating the representations from the \textit{race} attribute and (ii) bringing the representations with the same \textit{gender} closer to each other.  

\begin{table}[t]
\begin{center}
 \caption{Evaluation of the proposed method on UTKFace for 2 different protected attributes, namely \textit{race} and \textit{age}, with \textit{gender} as the target attribute.}
    \label{tab:utk}
 \begin{tabular}{lcccccc} 
 \toprule
 
         \multirow{ 3}{*}{methods} & \multirow{ 3}{*}{$L_{task}$}  &  \multirow{ 3}{*}{BL-U} & \multicolumn{4}{c}{bias}   \\ \cline{4-7}
        & & & \multicolumn{2}{c}{race} &  \multicolumn{2}{c}{age} \\ \cline{4-7}
        & & & unbiased & bias-conflict & unbiased & bias-conflict \\ \midrule
Vanilla & CE & \cmark& 87.4$\pm$0.3 & 79.1$\pm$0.3 & 72.3$\pm$0.3 & 46.5$\pm$0.2 \\ 
         LNL & CE & \xmark& 87.3$\pm$0.3  & 78.8$\pm$0.6 & 72.9$\pm$0.1  & 47.0$\pm$0.1 \\ 
         DI & CE & \xmark& 88.9$\pm$1.2 & 89.1$\pm$1.6 & 75.6$\pm$0.8  & 60.0$\pm$0.2 \\ 
         EnD & CE & \xmark& 88.4$\pm$0.3 & 81.6$\pm$0.3 & 73.2$\pm$0.3 & 47.9$\pm$0.6 \\ 
         BC-BB & Con. &\xmark & \dotuline{91.0}$\pm$0.2 & \dotuline{89.2}$\pm$0.1 & \dotuline{79.1}$\pm$0.3 & \dotuline{71.7}$\pm$0.8 \\ 
         FairKL\footnotemark & Con. & \xmark & 85.5$\pm$0.7 & 80.4$\pm$1.0 & 72.7$\pm$0.2  & 48.6$\pm$0.6 \\
         SoftCon & Con. & \cmark& 87.0 & 80.2 & 74.6 & 59.2 \\ \midrule
         \METHOD & CE &\cmark &  \underline{90.9}$\pm$0.3 & 90.2$\pm$0.4 & 78.7$\pm$1.3 & \textbf{81.1}$\pm$1.6 \\
        \METHOD & Con. &\cmark &  \textbf{92.0}$\pm$0.2 & \underline{92.2}$\pm$0.7 & \underline{80.6}$\pm$0.7 & 71.6$\pm$2.6 \\
        \METHOD-B & Con. &\cmark &  91.4$\pm$0.2 & \textbf{93.2}$\pm$0.3 & \textbf{80.7}$\pm$0.7 & \underline{77.0}$\pm$3.7 \\
        \bottomrule
\end{tabular} 
\end{center}
\end{table}
\footnotetext{Reimplemented based on the code provided by the authors.}

\begin{table}[H]
\begin{center}
\caption{Evaluation on Corrupted-Cifar10. 
Underlined and dotted-underlined values refer to the second best performing BL-U and the best performing BL-A methods, respectively. Con. refer to Contrastive losses.}\label{tab:cifar}
    \begin{tabular}{lcccccc} 
    \toprule
          \multirow{ 2}{*}{methods} & \multirow{ 2}{*}{$L_{task}$}  &  \multirow{ 2}{*}{BL-U} &    \multicolumn{4}{c}{$q$}     \\ \cline{4-7}
 & & & 0.95 & 0.98 & 0.99 & 0.995 \\ 
\midrule
 Vanilla & CE & \cmark & 39.4$\pm$0.6 & 30.1$\pm$0.7 & 25.8$\pm$0.3 & 23.1$\pm$1.2\\ 
         
          EnD& CE & \xmark&  36.6$\pm$4.0 & 34.1$\pm$4.8 &  23.1$\pm$1.1 & 19.4$\pm$1.4\\
         
         FairKL & Con. & \xmark& \dotuline{50.7}$\pm$0.9 & \dotuline{41.5}$\pm$0.4 & \dotuline{36.5}$\pm$0.4 & \dotuline{33.3}$\pm$0.4\\
         HEX & CE &\cmark&  16.0$\pm$0.6 & 15.2$\pm$0.5 & 14.8$\pm$0.4 & 13.9$\pm$0.0\\
         ReBias & CE &\cmark& 43.4$\pm$0.4 & 31.7$\pm$0.4 & 25.7$\pm$0.2 & 22.3$\pm$0.4\\
         LfF & CE &\cmark& {50.3}$\pm$1.6 & {39.9}$\pm$0.3 & {33.1}$\pm$0.8 & 28.6$\pm$1.3\\ 
         DisEnt& CE & \cmark&  \underline{51.1}$\pm$1.3 & \underline{41.8}$\pm$2.3 &  \underline{36.5}$\pm$1.8 & \underline{30.0}$\pm$0.7\\ \midrule
         
         \METHOD-B & Con. &\cmark& \textbf{53.0}$\pm$0.7  & \textbf{46.0}$\pm$0.2 & \textbf{39.3}$\pm$0.4 & \textbf{34.1}$\pm$0.5\\
         \bottomrule
         
\end{tabular}
\end{center}
\end{table}

\begin{table}[h]
\centering
 \caption{Evaluation of the proposed method on the biased and the unbiased 9-Class ImageNet test sets and the ImageNet-A. FairKL-lu refers to the BL-U version of FairKL.}
    \label{tab:imagenet}
    \begin{tabular}{lcccc} 
    \toprule
    methods & $\mathcal{L}_{task}$ & biased & unbiased & ImageNet-A \\ \midrule

         Vanilla & CE & 94.0$\pm$0.1 & 92.7$\pm$0.2& 30.5$\pm$0.5 \\ 
         LM & CE  &  79.2$\pm$1.1 & 76.6$\pm$1.2  &  19.0$\pm$1.2  \\
         Rubi & CE & 93.9$\pm$0.2  &  92.5$\pm$0.2 & 31.0$\pm$0.2   \\
         ReBias & CE  & 94.0$\pm$0.2 & 92.7$\pm$0.2  & 30.5$\pm$0.2   \\
         LfF & CE & 91.2$\pm$0.1  & 89.6$\pm$0.3  &  29.4$\pm$0.8  \\
         SoftCon & Con.  & 95.3$\pm$0.2  &  94.1$\pm$0.3 & 34.1$\pm$0.6   \\
         FairKL-lu & Con.  &  95.1$\pm$0.1 &  \underline{94.8}$\pm$0.3  &  35.7$\pm$0.5  \\ \midrule
         \METHOD-B & CE  & \underline{95.5}$\pm$0.2  & \textbf{95.2}$\pm$0.2 & \underline{37.6}$\pm$0.6\\
         \METHOD-B & Con.  & \textbf{95.7}$\pm$0.2  & \textbf{95.2}$\pm$0.2 & \textbf{37.9}$\pm$0.9\\
         \bottomrule
    \end{tabular}
\end{table}

The last evaluation scenario does not refer to societal biases, but biases that are introduced from the image's background or texture. 
The results on the Corrupted-Cifar10 for 4 different correlation ratios are presented in Table \ref{tab:cifar}. It is noteworthy that FLAC consistently outperforms the second best performing method for all the different correlation ratios, namely 2.9\%, 4.2\%, 2.8\%, and 4.1\% for $q$ equal to 0.95, 0.98, 0.99, and 0.995, respectively.
Table \ref{tab:imagenet} presents the results of the experiments conducted on the 9-Class ImageNet dataset. In this scenario, the bias is not categorical, thus only BL-U methods can be applied. As presented in Table \ref{tab:imagenet}, \METHOD\ outperforms the state-of-the-art on all three test sets. As in previous experiments, \METHOD\ demonstrates 2.2\% improvements in terms of accuracy on the most challenging test set (i.e., ImageNet-A).

\begin{figure}[H]
    \centering
    \includegraphics[width=0.7\linewidth,trim={0.6cm 0 1.6cm 0},clip]{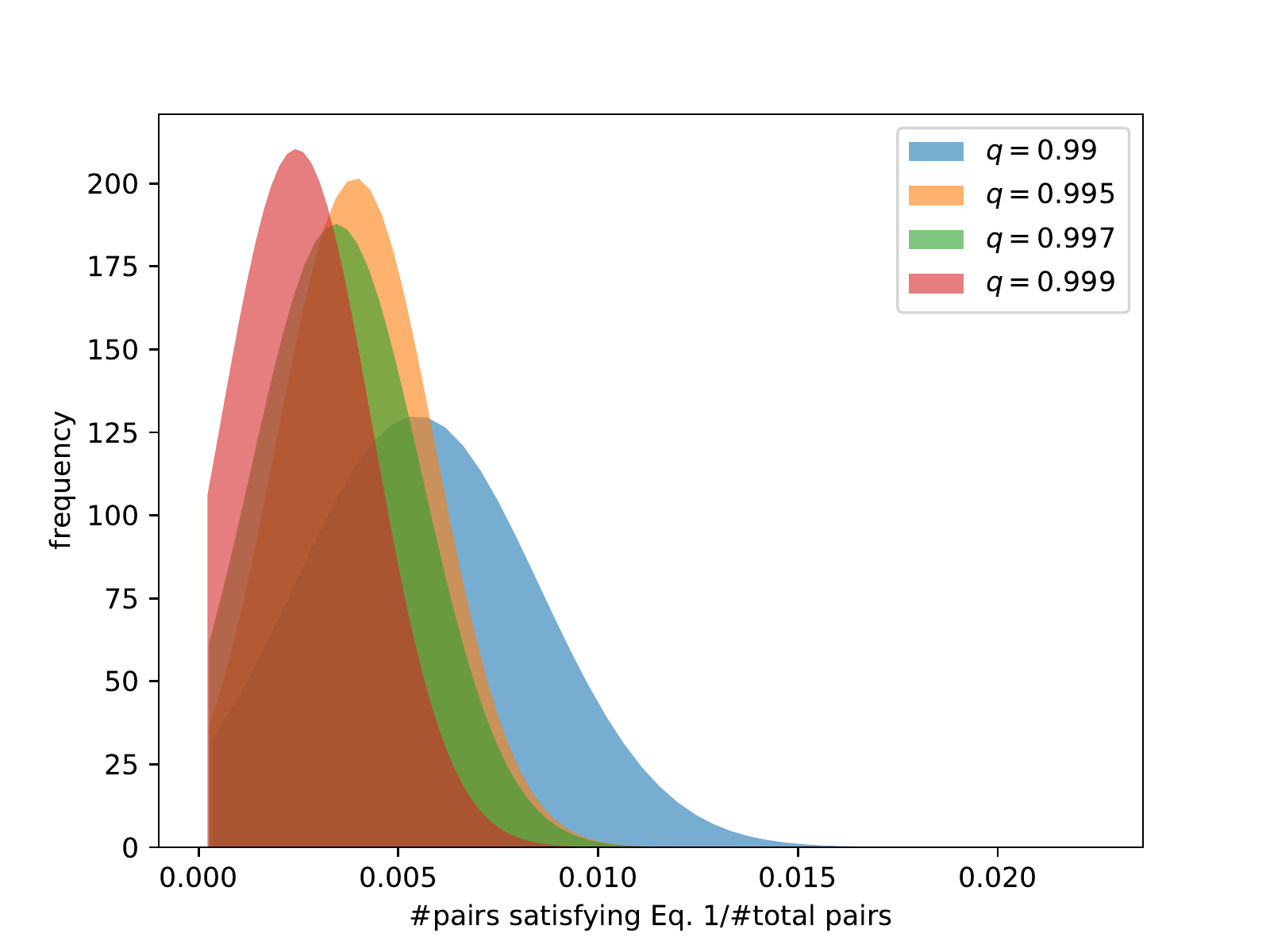}
    \caption{The frequency of the number of sample pairs belonging to $\mathcal{S}$ per batch ($N=128$) for Biased-MNIST training set with $p\in\{0.99,0.995,0.997,0.999\}$.} 
    \label{fig:num_of_pairs}
\end{figure}
\section{Ablation study}\label{sec:ablation}

\METHOD\ is designed to be applied only on the meaningful pairs of samples (see Section \ref{sec:analysis}), which is not the case for other BL-U approaches. Particularly, other works \cite{hong2021bb,barbano2022fairkl} involve all the possible sample pairs by assigning weights based on their similarities. However, the meaningful pairs of samples are only a very small subset of the set of all the possible pairs 
, especially in extremely biased data, such as the Biased-MNIST with $q=0.999$ or the CelebA with \textit{HeavyMakeup} as target. As presented in Figure \ref{fig:num_of_pairs}, the maximum number of sample pairs in the Biased-MNIST with $q=0.99$ that satisfy Equation \eqref{eq:cond} is 366 (i.e., only 0.022\% of all pairs), while the number of all the possible sample pairs is $128^2$. Involving all sample pairs in the bias mitigation procedure makes competing BL-U methods much more sensitive to scenarios where only few sample pairs in the dataset are responsible for the  emergence of bias. 
Table \ref{tab:eq1} compares the performance of \METHOD\ and two state-of-the-art BL-U approaches, namely SoftCon \cite{hong2021bb} and FairKL-lu \cite{barbano2022fairkl}, w/ and w/o leveraging the sampling based on Equation \eqref{eq:cond}. It becomes obvious that using Equation \eqref{eq:cond} can significantly improve the performance of other  BL-U methods, while \METHOD\ w/ Equation \eqref{eq:cond} still exhibits the best performance. Furthermore, the importance of involving both terms of Equation \eqref{eq:cond} as analyzed in Section \ref{sec:analysis} is experimentally confirmed in Table \ref{tab:cond_terms}, which presents the performance of \METHOD\ using only the first (i.e., pairs with the same target and different protected attribute label) or the second  term (i.e., pairs with different target and same protected attribute label) of Equation \eqref{eq:cond}. 

\begin{table}[h]
\centering
    
\caption{Performance of state-of-the-art BL-U methods w/o or w/ adopting Equation \eqref{eq:cond}. Results pertain to BiasedMNIST with $q=0.999$.}\label{tab:eq1}

\centering
\begin{tabular}{lccc}\toprule%
method & $\mathcal{L}_{task}$ & acc. w/o Eq. \eqref{eq:cond} & acc. w/ Eq. \eqref{eq:cond} \\ \midrule
SoftCon & Con. & 65.0 & 84.0 \\
FairKL-lu  & Con. & 13.7 & 73.9 \\ \midrule
\METHOD\ & CE &  24.1 & 89.3\\
\METHOD\ & Con. &  28.2 & \textbf{94.1}\\
\bottomrule
\end{tabular}
\end{table}

\begin{table}[h]
\centering
\caption{The impact of the two terms of Equation \eqref{eq:cond} on the performance of \METHOD. Results pertain to BiasedMNIST with $q=0.999$.}
    \label{tab:cond_terms}
     \centering
\begin{tabular}{ccc}\toprule%
 Equation \eqref{eq:cond} variants & acc. (CE) & acc. (Con.) \\ \midrule

 $y_i \neq y_j \land t_i = t_j  $  & 10.1 & 29.1\\
    $y_i = y_j \land t_i \neq t_j $  & 10.5 & 20.6 \\
     $(y_i = y_j \land t_i \neq t_j) \lor (y_i \neq y_j \land t_i = t_j)$ & \textbf{89.3} & \textbf{94.1} \\ 
\bottomrule
\end{tabular}
\end{table}

\begin{table}[h]
\centering
\caption{The impact of kernel function on the performance of \METHOD. Results pertain to BiasedMNIST with $q=0.999$.}
    \label{tab:kernels}
     \centering
\begin{tabular}{ccc} \toprule
kernel & accuracy (CE) & accuracy (Con.) \\ \midrule
cosine   & 60.4 & 82.8 \\
     RBF   & 83.4 & 90.7 \\
     student's t & \textbf{89.3} & \textbf{94.1}  \\  \bottomrule
\end{tabular}
\end{table}

\begin{table}[h]

\centering
\caption{Performance of \METHOD\ with MSE, Kullback-Leibler, and Jeffreys divergence as loss functions. Results pertain to BiasedMNIST with $q=0.999$.}
    \label{tab:losses}
     \centering
\begin{tabular}{ccc}
\toprule%
loss & accuracy (CE) & accuracy (Con.) \\ \midrule
 MSE  & 12.1 & 85.6 \\ 
        Kullback-Leibler divergence  & 85.2 & 89.2 \\
        Jeffreys divergence (Eq. \eqref{eq:jef})  & \textbf{89.3} & \textbf{94.1}\\ 
\bottomrule
\end{tabular}

\end{table}

\begin{table}[h!]
\centering
 \caption{The impact of \METHOD\ when the training data is unbiased. Results for Biased-MNIST with $q=0.1$.}
    \label{tab:corr01}
     \centering
\begin{tabular}{ccc}\toprule%
method  & accuracy (CE) & accuracy (Con.) \\
\midrule
Vanilla    & 99.3 & -\\
     \METHOD@$\alpha$=1   & 99.4 & 99.3\\
    \METHOD@$\alpha$=100  & 99.1 & 99.1\\ 
    \METHOD@$\alpha$=1000 &  98.2 &  98.9 \\
\bottomrule
\end{tabular}

\end{table}

Table \ref{tab:kernels} presents the impact of several kernel functions on the performance of the proposed method. The kernel function should be carefully selected as it can significantly affect the model's performance. For instance, although the cosine kernel fits well in retrieval tasks, it is not effective enough on classification tasks as previous studies have noticed \cite{passalis2020heterogeneous}, which is also reflected in Table \ref{tab:kernels}.  Furthermore, Radial Basis Function (RBF) kernels constitute a common kernel option, but tuning them is often difficult. Finally, the student's t kernel, which is considered as a good option for classification tasks, is shown to achieve the best performance compared to other kernels in the conducted experiments.

Table \ref{tab:losses} presents the evaluation of \METHOD\ using different loss functions, namely the MSE, Kullback-Leibler divergence, and Jeffreys divergence. As expected, using a common regression loss, such as MSE, is not a good choice to mitigate bias, while Kullback-Leibler and Jeffreys divergences lead to highly accurate models, with Jeffreys divergence leading to the best results.

In order to investigate whether the proposed method can have a negative impact on the model's performance when the training data is unbiased, we further evaluate the proposed method on a dataset that does not suffer from any attribute-label associations (i.e., Biased-MNIST with $q=0.1$).  Table \ref{tab:corr01} presents the performance of \METHOD\ for $\alpha \in \{1,100,1000\}$: for $\alpha=1$ and $\alpha=100$ \METHOD\ does not affect the model's performance, while for the extreme value $\alpha=1000$ only a small drop in accuracy is noticed.

\section{Conclusion}
In this paper, we introduce  \METHOD, a bias mitigation approach that leverages the representations of a bias-capturing classifier for enabling the main model to learn fair representations without being aware of the protected attribute labels. In particular, \METHOD\ aims to minimize the MI between the main model's representations and the protected attribute without taking advantage of the protected attribute labels. To this end, \METHOD\ only leverages the under-represented samples that can effectively contribute to minimizing MI. By doing so, the derived fair representations do not capture any information related to the protected attributes. The proposed approach demonstrates superior performance compared to state-of-the-art in a wide range of experiments on both datasets with artificially injected bias and established computer vision datasets. As regards the limitations of the proposed approach, it should be highlighted that FLAC necessitates the presence of some bias-conflicting samples in the training data and thus cannot be applied in 100\% correlation scenarios. Furthermore, its effectiveness diminishes for batch sizes equal to or lower than 32, and it does not account for multiple sources of bias. Thus, as a future work, we consider enabling \METHOD\ to be applied in a multi-attribute fairness scenario. Furthermore, taking into account that bias is uniformly distributed across classes in the existing fairness benchmarks, another future work could be the exploration of fairness-aware approaches performance on datasets with non-uniform bias distribution.  Finally, using \METHOD\ for tasks beyond classification ( e.g., retrieval), is a potential subject for future work. 

\section*{Acknowledgments}
This research was supported by the EU Horizon Europe projects
MAMMOth (Grant Agreement 101070285), ELIAS (Grant Agreement 101120237), and the EU H2020 project  MediaVerse (Grant
Agreement 957252).

\bibliographystyle{splncs04}
\bibliography{references}  

\begin{thebibliography}{10}
\providecommand{\url}[1]{\texttt{#1}}
\providecommand{\urlprefix}{URL }
\providecommand{\doi}[1]{https://doi.org/#1}

\bibitem{adel2019one}
Adel, T., Valera, I., Ghahramani, Z., Weller, A.: One-network adversarial
  fairness. In: Proceedings of the AAAI Conference on Artificial Intelligence.
  vol.~33, pp. 2412--2420 (2019)

\bibitem{ali1966general}
Ali, S.M., Silvey, S.D.: A general class of coefficients of divergence of one
  distribution from another. Journal of the Royal Statistical Society: Series B
  (Methodological)  \textbf{28}(1),  131--142 (1966)

\bibitem{alvi2018turning}
Alvi, M., Zisserman, A., Nell{\aa}ker, C.: Turning a blind eye: Explicit
  removal of biases and variation from deep neural network embeddings. In:
  Proceedings of the European Conference on Computer Vision (ECCV) Workshops.
  pp.~0--0 (2018)

\bibitem{bahng2020rebias}
Bahng, H., Chun, S., Yun, S., Choo, J., Oh, S.J.: Learning de-biased
  representations with biased representations. In: International Conference on
  Machine Learning. pp. 528--539. PMLR (2020)

\bibitem{barbano2022fairkl}
Barbano, C.A., Dufumier, B., Tartaglione, E., Grangetto, M., Gori, P.: Unbiased
  supervised contrastive learning. arXiv preprint arXiv:2211.05568  (2022)

\bibitem{barocasfairml}
Barocas, S., Hardt, M., Narayanan, A.: Fairness and Machine Learning:
  Limitations and Opportunities. fairmlbook.org (2019),
  \url{http://www.fairmlbook.org}

\bibitem{bobadilla2013recommender}
Bobadilla, J., Ortega, F., Hernando, A., Guti{\'e}rrez, A.: Recommender systems
  survey. Knowledge-based systems  \textbf{46},  109--132 (2013)

\bibitem{boudiaf2020unifying}
Boudiaf, M., Rony, J., Ziko, I.M., Granger, E., Pedersoli, M., Piantanida, P.,
  Ayed, I.B.: A unifying mutual information view of metric learning:
  cross-entropy vs. pairwise losses. In: Computer Vision--ECCV 2020: 16th
  European Conference, Glasgow, UK, August 23--28, 2020, Proceedings, Part VI.
  pp. 548--564. Springer (2020)

\bibitem{brendel2019approximating}
Brendel, W., Bethge, M.: Approximating cnns with bag-of-local-features models
  works surprisingly well on imagenet. arXiv preprint arXiv:1904.00760  (2019)

\bibitem{cadene2019rubi}
Cadene, R., Dancette, C., Cord, M., Parikh, D., et~al.: Rubi: Reducing unimodal
  biases for visual question answering. Advances in neural information
  processing systems  \textbf{32} (2019)

\bibitem{cavazos2020accuracy}
Cavazos, J.G., Phillips, P.J., Castillo, C.D., O’Toole, A.J.: Accuracy
  comparison across face recognition algorithms: Where are we on measuring race
  bias? IEEE transactions on biometrics, behavior, and identity science
  \textbf{3}(1),  101--111 (2020)

\bibitem{clark2019LM}
Clark, C., Yatskar, M., Zettlemoyer, L.: Don’t take the easy way out:
  Ensemble based methods for avoiding known dataset biases. In: Proceedings of
  the 2019 Conference on Empirical Methods in Natural Language Processing and
  the 9th International Joint Conference on Natural Language Processing
  (EMNLP-IJCNLP). pp. 4069--4082 (2019)

\bibitem{creswell2018generative}
Creswell, A., White, T., Dumoulin, V., Arulkumaran, K., Sengupta, B., Bharath,
  A.A.: Generative adversarial networks: An overview. IEEE signal processing
  magazine  \textbf{35}(1),  53--65 (2018)

\bibitem{das2019separability}
Das, R., Chaudhuri, S.: On the separability of classes with the cross-entropy
  loss function. arXiv preprint arXiv:1909.06930  (2019)

\bibitem{deng2009imagenet}
Deng, J., Dong, W., Socher, R., Li, L.J., Li, K., Fei-Fei, L.: Imagenet: A
  large-scale hierarchical image database. In: 2009 IEEE Conference on Computer
  Vision and Pattern Recognition. pp. 248--255. IEEE (2009)

\bibitem{fabbrizzi2022survey}
Fabbrizzi, S., Papadopoulos, S., Ntoutsi, E., Kompatsiaris, I.: A survey on
  bias in visual datasets. Computer Vision and Image Understanding
  \textbf{223},  103552 (2022)

\bibitem{frisella2022quantifying}
Frisella, M., Khorrami, P., Matterer, J., Kratkiewicz, K., Torres-Carrasquillo,
  P.: Quantifying bias in a face verification system. In: Computer Sciences \&
  Mathematics Forum. vol.~3, p.~6. MDPI (2022)

\bibitem{gluge2020not}
Gl{\"u}ge, S., Amirian, M., Flumini, D., Stadelmann, T.: How (not) to measure
  bias in face recognition networks. In: IAPR Workshop on Artificial Neural
  Networks in Pattern Recognition. pp. 125--137. Springer (2020)

\bibitem{gretton2005measuring}
Gretton, A., Bousquet, O., Smola, A., Sch{\"o}lkopf, B.: Measuring statistical
  dependence with hilbert-schmidt norms. In: Algorithmic Learning Theory: 16th
  International Conference, ALT 2005, Singapore, October 8-11, 2005.
  Proceedings 16. pp. 63--77. Springer (2005)

\bibitem{gupta2018robot}
Gupta, A., Murali, A., Gandhi, D.P., Pinto, L.: Robot learning in homes:
  Improving generalization and reducing dataset bias. Advances in neural
  information processing systems  \textbf{31} (2018)

\bibitem{he2016deep}
He, K., Zhang, X., Ren, S., Sun, J.: Deep residual learning for image
  recognition. In: Proceedings of the IEEE conference on computer vision and
  pattern recognition. pp. 770--778 (2016)

\bibitem{hendrycks2018benchmarking}
Hendrycks, D., Dietterich, T.G.: Benchmarking neural network robustness to
  common corruptions and surface variations. arXiv preprint arXiv:1807.01697
  (2018)

\bibitem{hendrycks2021nae}
Hendrycks, D., Zhao, K., Basart, S., Steinhardt, J., Song, D.: Natural
  adversarial examples. CVPR  (2021)

\bibitem{hong2021bb}
Hong, Y., Yang, E.: Unbiased classification through bias-contrastive and
  bias-balanced learning. Advances in Neural Information Processing Systems
  \textbf{34},  26449--26461 (2021)

\bibitem{janssen2021bias}
Janssen, P., Sadowski, B.M.: Bias in algorithms: On the trade-off between
  accuracy and fairness  (2021)

\bibitem{jung2022learning}
Jung, S., Chun, S., Moon, T.: Learning fair classifiers with partially
  annotated group labels. In: Proceedings of the IEEE/CVF Conference on
  Computer Vision and Pattern Recognition. pp. 10348--10357 (2022)

\bibitem{khosla2020supervised}
Khosla, P., Teterwak, P., Wang, C., Sarna, A., Tian, Y., Isola, P., Maschinot,
  A., Liu, C., Krishnan, D.: Supervised contrastive learning. Advances in
  neural information processing systems  \textbf{33},  18661--18673 (2020)

\bibitem{kim2019lnl}
Kim, B., Kim, H., Kim, K., Kim, S., Kim, J.: Learning not to learn: Training
  deep neural networks with biased data. In: Proceedings of the IEEE/CVF
  Conference on Computer Vision and Pattern Recognition. pp. 9012--9020 (2019)

\bibitem{krasanakis2018adaptive}
Krasanakis, E., Spyromitros-Xioufis, E., Papadopoulos, S., Kompatsiaris, Y.:
  Adaptive sensitive reweighting to mitigate bias in fairness-aware
  classification. In: Proceedings of the 2018 world wide web conference. pp.
  853--862 (2018)

\bibitem{kraskov2005hierarchical}
Kraskov, A., St{\"o}gbauer, H., Andrzejak, R.G., Grassberger, P.: Hierarchical
  clustering using mutual information. Europhysics Letters  \textbf{70}(2),
  ~278 (2005)

\bibitem{lecun1998mnist}
LeCun, Y.: The mnist database of handwritten digits. http://yann. lecun.
  com/exdb/mnist/  (1998)

\bibitem{lee2021learning}
Lee, J., Kim, E., Lee, J., Lee, J., Choo, J.: Learning debiased representation
  via disentangled feature augmentation. Advances in Neural Information
  Processing Systems  \textbf{34},  25123--25133 (2021)

\bibitem{lee2023revisiting}
Lee, J., Park, J., Kim, D., Lee, J., Choi, E., Choo, J.: Revisiting the
  importance of amplifying bias for debiasing. In: Proceedings of the AAAI
  Conference on Artificial Intelligence. vol.~37, pp. 14974--14981 (2023)

\bibitem{li2021multi}
Li, W., Liang, Z., Neuman, J., Chen, J., Cui, X.: Multi-generator gan learning
  disconnected manifolds with mutual information. Knowledge-Based Systems
  \textbf{212},  106513 (2021)

\bibitem{liu2022reducing}
Liu, D., Qu, X., Hu, W.: Reducing the vision and language bias for temporal
  sentence grounding. In: Proceedings of the 30th ACM International Conference
  on Multimedia. pp. 4092--4101 (2022)

\bibitem{liu2015faceattributes}
Liu, Z., Luo, P., Wang, X., Tang, X.: Deep learning face attributes in the
  wild. In: Proceedings of International Conference on Computer Vision (ICCV)
  (December 2015)

\bibitem{lum2016statistical}
Lum, K., Johndrow, J.: A statistical framework for fair predictive algorithms.
  arXiv preprint arXiv:1610.08077  (2016)

\bibitem{meier1984happened}
Meier, P., Sacks, J., Zabell, S.L.: What happened in hazelwood: Statistics,
  employment discrimination, and the 80\% rule. American Bar Foundation
  Research Journal  \textbf{9}(1),  139--186 (1984)

\bibitem{nam2020LfF}
Nam, J., Cha, H., Ahn, S., Lee, J., Shin, J.: Learning from failure: De-biasing
  classifier from biased classifier. Advances in Neural Information Processing
  Systems  \textbf{33},  20673--20684 (2020)

\bibitem{nam2022spread}
Nam, J., Kim, J., Lee, J., Shin, J.: Spread spurious attribute: Improving
  worst-group accuracy with spurious attribute estimation. In: International
  Conference on Learning Representations (2022)

\bibitem{ntoutsi2020bias}
Ntoutsi, E., Fafalios, P., Gadiraju, U., Iosifidis, V., Nejdl, W., Vidal, M.E.,
  Ruggieri, S., Turini, F., Papadopoulos, S., Krasanakis, E., et~al.: Bias in
  data-driven artificial intelligence systems—an introductory survey. Wiley
  Interdisciplinary Reviews: Data Mining and Knowledge Discovery
  \textbf{10}(3),  e1356 (2020)

\bibitem{passalis2018learning}
Passalis, N., Tefas, A.: Learning deep representations with probabilistic
  knowledge transfer. In: Proceedings of the European Conference on Computer
  Vision (ECCV). pp. 268--284 (2018)

\bibitem{passalis2020heterogeneous}
Passalis, N., Tzelepi, M., Tefas, A.: Heterogeneous knowledge distillation
  using information flow modeling. In: Proceedings of the IEEE/CVF Conference
  on Computer Vision and Pattern Recognition. pp. 2339--2348 (2020)

\bibitem{sarridis2022indistill}
Sarridis, I., Koutlis, C., Papadopoulos, S., Kompatsiaris, I.: Indistill:
  Transferring knowledge from pruned intermediate layers. arXiv preprint
  arXiv:2205.10003  (2022)

\bibitem{sattigeri2018fairness}
Sattigeri, P., Hoffman, S.C., Chenthamarakshan, V., Varshney, K.R.: Fairness
  gan. stat  \textbf{1050}, ~24 (2018)

\bibitem{sixta2020fairface}
Sixta, T., Jacques~Junior, J., Buch-Cardona, P., Vazquez, E., Escalera, S.:
  Fairface challenge at eccv 2020: Analyzing bias in face recognition. In:
  European conference on computer vision. pp. 463--481. Springer (2020)

\bibitem{song2019learning}
Song, J., Kalluri, P., Grover, A., Zhao, S., Ermon, S.: Learning controllable
  fair representations. In: The 22nd International Conference on Artificial
  Intelligence and Statistics. pp. 2164--2173. PMLR (2019)

\bibitem{taigman2014deepface}
Taigman, Y., Yang, M., Ranzato, M., Wolf, L.: Deepface: Closing the gap to
  human-level performance in face verification. In: Proceedings of the IEEE
  conference on computer vision and pattern recognition. pp. 1701--1708 (2014)

\bibitem{tan2020efficientdet}
Tan, M., Pang, R., Le, Q.V.: Efficientdet: Scalable and efficient object
  detection. In: Proceedings of the IEEE/CVF conference on computer vision and
  pattern recognition. pp. 10781--10790 (2020)

\bibitem{tartaglione2021end}
Tartaglione, E., Barbano, C.A., Grangetto, M.: End: Entangling and
  disentangling deep representations for bias correction. In: Proceedings of
  the IEEE/CVF conference on computer vision and pattern recognition. pp.
  13508--13517 (2021)

\bibitem{torkkola2003feature}
Torkkola, K.: Feature extraction by non-parametric mutual information
  maximization. Journal of machine learning research  \textbf{3}(Mar),
  1415--1438 (2003)

\bibitem{vergara2014review}
Vergara, J.R., Est{\'e}vez, P.A.: A review of feature selection methods based
  on mutual information. Neural computing and applications  \textbf{24},
  175--186 (2014)

\bibitem{wang2019balanced}
Wang, T., Zhao, J., Yatskar, M., Chang, K.W., Ordonez, V.: Balanced datasets
  are not enough: Estimating and mitigating gender bias in deep image
  representations. In: Proceedings of the IEEE/CVF International Conference on
  Computer Vision. pp. 5310--5319 (2019)

\bibitem{wang2020DI}
Wang, Z., Qinami, K., Karakozis, I.C., Genova, K., Nair, P., Hata, K.,
  Russakovsky, O.: Towards fairness in visual recognition: Effective strategies
  for bias mitigation. In: Proceedings of the IEEE/CVF conference on computer
  vision and pattern recognition. pp. 8919--8928 (2020)

\bibitem{xie2017controllable}
Xie, Q., Dai, Z., Du, Y., Hovy, E., Neubig, G.: Controllable invariance through
  adversarial feature learning. Advances in neural information processing
  systems  \textbf{30} (2017)

\bibitem{xu2018fairgan}
Xu, D., Yuan, S., Zhang, L., Wu, X.: Fairgan: Fairness-aware generative
  adversarial networks. In: 2018 IEEE International Conference on Big Data (Big
  Data). pp. 570--575. IEEE (2018)

\bibitem{xu2020investigating}
Xu, T., White, J., Kalkan, S., Gunes, H.: Investigating bias and fairness in
  facial expression recognition. In: Proceedings of the European Conference on
  Computer Vision (ECCV) Workshops. pp. 506--523 (2020)

\bibitem{yan2020mitigating}
Yan, S., Huang, D., Soleymani, M.: Mitigating biases in multimodal personality
  assessment. In: Proceedings of the 2020 International Conference on
  Multimodal Interaction. pp. 361--369 (2020)

\bibitem{zhao2022inherent}
Zhao, H., Gordon, G.J.: Inherent tradeoffs in learning fair representations.
  Journal of Machine Learning Research  \textbf{23},  1--26 (2022)

\bibitem{zheng2020survey}
Zheng, X., Guo, Y., Huang, H., Li, Y., He, R.: A survey of deep facial
  attribute analysis. International Journal of Computer Vision  \textbf{128},
  2002--2034 (2020)

\bibitem{zhifei2017cvpr}
Zhifei, Z., Yang, S., Hairong, Q.: Age progression/regression by conditional
  adversarial autoencoder. In: IEEE Conference on Computer Vision and Pattern
  Recognition (CVPR). IEEE (2017)

\end{thebibliography}
\newpage
\appendix
\section{Supplementary Material}
\label{metrics}
\METHOD\ is also evaluated in terms of three fairness metrics, namely \textit{p\% rule}, DFPR, and DFNR. \textit{p\% rule} is defined as: 
\begin{equation}
    \min \left \{ \frac{P(\hat{y_i}=1\mid t_i=0)}{P(\hat{y_i}=1\mid t_i=1)}, \frac{P(\hat{y_i}=1\mid t_i=1)}{P(\hat{y_i}=1\mid t_i=0)} \right \} \geq \frac{p}{100},
\end{equation}
where $p$\% should be at least 80\% to consider the model's outcomes fair. DFPR and DFNR are defined as follows: 
\begin{equation}
\begin{split}
   D_{FPR} & = P\left(\hat{y}_i \neq y_i \mid y_i=0, t_i=1 \right)  -P\left(\hat{y}_i \neq y_i \mid y_i=0, t_i=0 \right), 
\end{split}
\end{equation}
\begin{equation}
\begin{split}
   D_{FNR} & = P\left(\hat{y}_i \neq y_i \mid y_i=1, t_i=1 \right) -P\left(\hat{y}_i \neq y_i \mid y_i=1, t_i=0 \right).
\end{split}
\end{equation}
Typically, these two metrics are combined into the following quantity to measure the overall disparate mistreatment: 
\begin{equation}
    \mid D_{FPR} \mid + \mid D_{FNR} \mid.
\end{equation}
\begin{table}[h]
    \centering
    \caption{Performance of \METHOD\ compared to Vanilla in terms of \textit{p\% rule} and $\mid D_{FPR} \mid + \mid D_{FNR}\mid$ on UTKFace and CelabA datasets.}
    \begin{tabular}{lcccc}
    \toprule
        dataset & method & $\mathcal{L}_{task}$ & \textit{p\% rule} & $\mid D_{FPR} \mid + \mid D_{FNR}\mid$ \\ \midrule
        \multirow{3}{*}{UTKFace (race)} & Vanilla & CE & 0.736   & 0.255 \\
        & \METHOD & CE & 0.985  & \textbf{0.013} \\
        & \METHOD & Con. & \textbf{0.998}  & 0.063 \\\midrule
        \multirow{3}{*}{UTKFace (age)} 
        &  Vanilla & CE & 0.450   & 0.447   \\
        & \METHOD & CE & \textbf{0.803}  & 0.424 \\
        & \METHOD & Con. & 0.759  & \textbf{0.419} \\\midrule
        \multirow{2}{*}{CelebA (BlondHair)} 
        &  Vanilla & CE &  0.324  & 0.526 \\
        & \METHOD & CE & \textbf{0.897} & \textbf{0.078} \\ & \METHOD & Con. & 0.890 & 0.093 \\ \midrule
        \multirow{2}{*}{CelebA (HeavyMakeup)} 
        &  Vanilla & CE &  0.250  & \textbf{0.113} \\
        & \METHOD & CE & 0.800  & 0.225 \\
        & \METHOD & Con. & \textbf{0.889}  & 0.344 \\
        \bottomrule
    \end{tabular} 
    
    \label{tab:prule}
\end{table}
In Table \ref{tab:prule}, we compare the performance of Vanilla and \METHOD\ in terms of these fairness metrics. As it can be noticed, not only \METHOD\ demonstrates higher \textit{p\%} values, but it also satisfies the lower bound of 80\% in all cases. This indicates that \METHOD\ results in fair representations for both UTKFace (with \textit{race} and \textit{gender} as protected attributes) and CelebA (with \textit{BlondHair} and \textit{HeavyMakeup} as targets). Furthermore, \METHOD\ demonstrates reduced overall disparate mistreatment compared to Vanilla in all experiments except for CelebA (HeavyMakeup). Note that occasionally higher $\mid D_{FPR} \mid$ and $\mid D_{FNR} \mid$ values can be attributed to the frequently encountered conflict between disparate mistreatment (as captured by these two metrics) and disparate impact ( measured by \textit{p\% rule}) mitigation.

\end{document}